\documentclass{article}

\usepackage{algorithm}
\usepackage[final]{corl_2026}

\usepackage{amsmath,amssymb,bm}
\usepackage{siunitx}
\usepackage{booktabs}
\usepackage{enumitem}
\usepackage{xurl}%
\usepackage{graphicx}
\graphicspath{{figures/}}
\usepackage[table]{xcolor}
\usepackage{tikz}
\usetikzlibrary{arrows.meta, positioning, fit, fadings, calc}
\tikzfading[name=wmrfade, left color=transparent!30, right color=transparent!80]
\tikzset{separator/.style={gray!55, dashed, semithick}}
\usepackage{pgfplots}
\pgfplotsset{compat=1.18}
\newcommand{\plotequal}{0.415\textwidth}
\newcommand{\plotgap}{0.09\textwidth}
\newcommand{\plotfull}{\dimexpr\plotequal+\plotgap+\plotequal\relax}
\newcommand{\plotthirdgap}{7mm}
\newcommand{\plotlegendcol}{21mm}
\newcommand{\plotthird}{\dimexpr(\plotfull-\plotthirdgap-\plotthirdgap-\plotlegendcol)/3\relax}
\newcommand{\plottall}{26mm}
\newcommand{\plotshort}{18.5mm}
\newcommand{\plotsep}{6.5mm}
\newcommand{\plothalf}{\dimexpr(\plottall-\plotsep)/2\relax}
\pgfplotsset{
  smplot/.style={
    scale only axis, axis lines*=left,
    axis line style=context, tick style=context, tick align=outside,
    label style={font=\footnotesize},
    tick label style={font=\footnotesize, /pgf/number format/assume math mode=true},
    title style={font=\footnotesize, at={(0,1)}, anchor=south west, inner xsep=0pt, execute at begin node=\strut}, every axis title shift=0pt,
    every axis plot/.append style={mark size=1.3pt, line join=round},
    legend style={draw=none, fill=none}, legend cell align=left,
    table/col sep=comma,
  },
  log number format basis/.code 2 args={#1\textsuperscript{\pgfmathprintnumber{#2}}},
  return axis/.style={ymin=0, ymax=0.9, ytick={0,0.4,0.8}},
}
\usepackage{algorithmicx,algpseudocode}
\definecolor{mydarkblue}{rgb}{0,0.08,0.45}
\usepackage{pifont}
\newcommand{\cmarkg}{\textcolor{green!55!black}{\checkmark}}
\newcommand{\xmarkr}{\textcolor{red!75!black}{\ding{55}}}

\definecolor{BerkeleyBlue}{HTML}{003262}
\definecolor{retcol}{RGB}{49,109,180}
\definecolor{AnaFill}{RGB}{217,240,211}

\definecolor{ours}{HTML}{084FA3}
\definecolor{theirs}{HTML}{BC5A29}
\definecolor{theirslight}{HTML}{D37F40}
\definecolor{theirsdark}{HTML}{9B411E}
\definecolor{context}{HTML}{595959}
\definecolor{track}{HTML}{D0D0D0}
\definecolor{region}{HTML}{F2F2F2}
\definecolor{highlight}{HTML}{DCE6F2}
\definecolor{shade}{HTML}{7CC79A}%

\newcommand{\frozenicon}[1][0.35cm]{\raisebox{-0.15ex}{\includegraphics[height=#1]{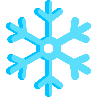}}}
\newcommand{\trainicon}[1][0.35cm]{\raisebox{-0.15ex}{\includegraphics[height=#1]{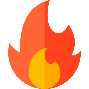}}}

\makeatletter
\renewcommand{\section}{%
  \@startsection{section}{1}{\z@}%
                {-1.5ex \@plus -0.4ex \@minus -0.2ex}%
                { 1.0ex \@plus  0.2ex \@minus  0.2ex}%
                {\large\bf\raggedright}%
}
\renewcommand{\subsection}{%
  \@startsection{subsection}{2}{\z@}%
                {-1.3ex \@plus -0.4ex \@minus -0.2ex}%
                { 0.6ex \@plus  0.2ex}%
                {\normalsize\bf\raggedright}%
}
\makeatother

\title{SuperMimic: Scaling Humanoid Whole-Body Control \\ with One Differentiable World Model}

\author{
  \\
  \vspace{-3em}
  \\
  Qiayuan Liao \quad
  Yiyang Shao$^{*}$ \quad
  Junfeng Long$^{*}$ \quad
  Kevin Zakka \quad \\
  Roberto Horowitz \quad 
  Pieter Abbeel$^{\dagger}$ \quad 
  Koushil Sreenath$^{\dagger}$
  \\
  \vspace{-0.5em}
  \\
  \textbf{UC Berkeley}
  \\
  \url{https://supermimic.github.io}
  \vspace{-3em}
}

\begin{document}
\maketitle
{\makeatletter\let\@footnotetext\H@@footnotetext\makeatother\renewcommand{\thefootnote}{*}\footnotetext{Equal contribution. \quad $^{\dagger}$Equal advising.}}

\begin{figure}[htbp]
\centering
\begin{minipage}[c]{0.65\textwidth}
  \centering
  \begin{tikzpicture}
    \node[inner sep=0pt] (timg)
      {\includegraphics[width=\linewidth, trim={150 75 120 125}, clip]{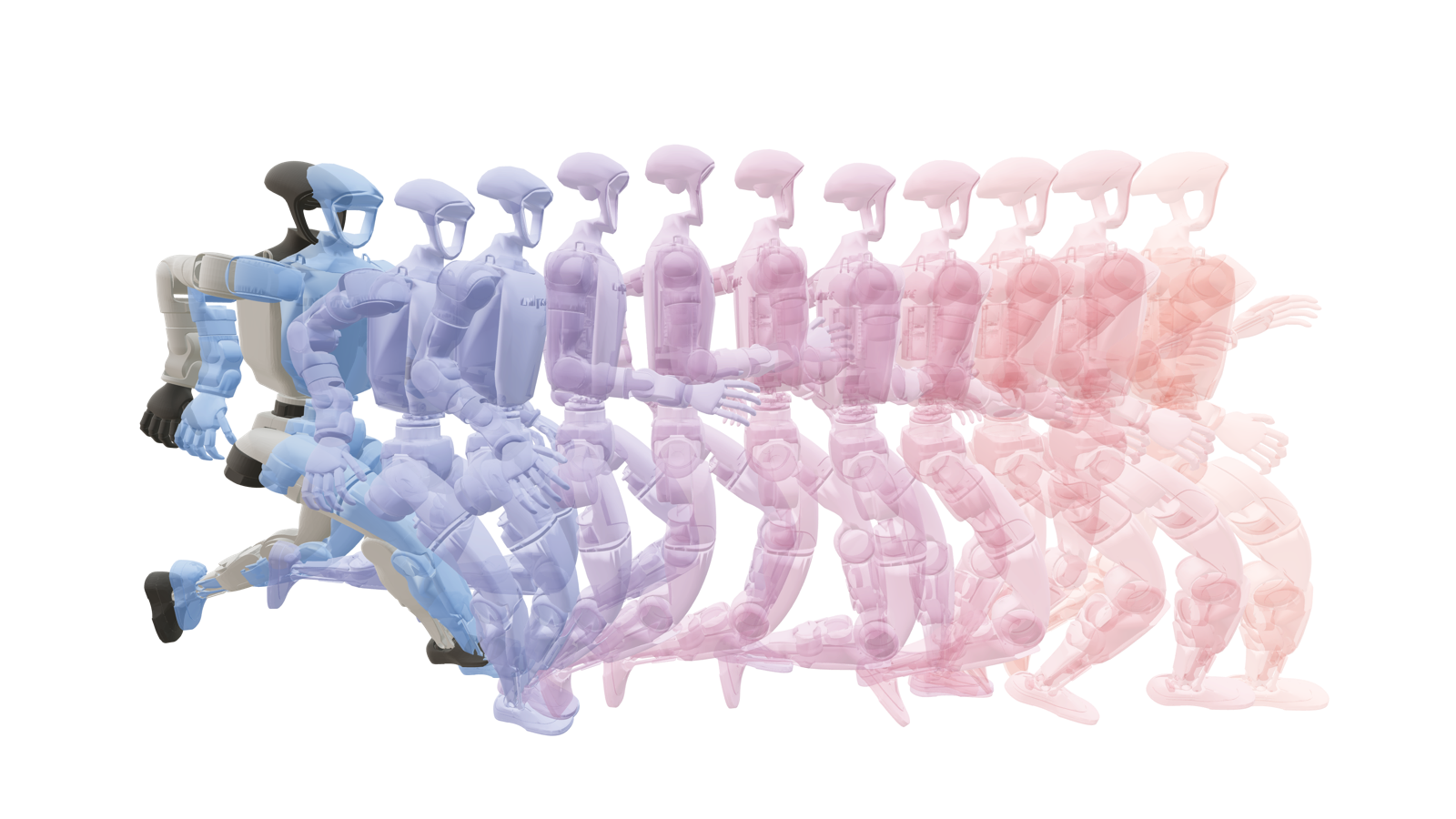}};
    \node[font=\large\bfseries, inner sep=3pt, fill=BerkeleyBlue,
          path fading=wmrfade, rounded corners=2pt, text=white]
      at (timg.center) {World-Model Rollout};
  \end{tikzpicture}
\end{minipage}
\tikz[baseline=0pt]\draw[separator] (0,-1.8) -- (0,1.8);
\begin{minipage}[c]{0.3\textwidth}
  \begin{tikzpicture}[font=\small]
    \pgfmathsetlengthmacro{\cellsz}{1.3*\linewidth/8}%
    \pgfmathsetlengthmacro{\cellp}{(\linewidth-\cellsz-4pt)/31}%
    \pgfmathsetlengthmacro{\cellpY}{1.12*\cellsz}%
    \foreach \r in {0,...,3}
      \foreach \c in {31,30,...,0}
        \node[inner sep=0pt] (s-\r-\c) at (\c*\cellp,\r*\cellpY) {\includegraphics[width=\cellsz]{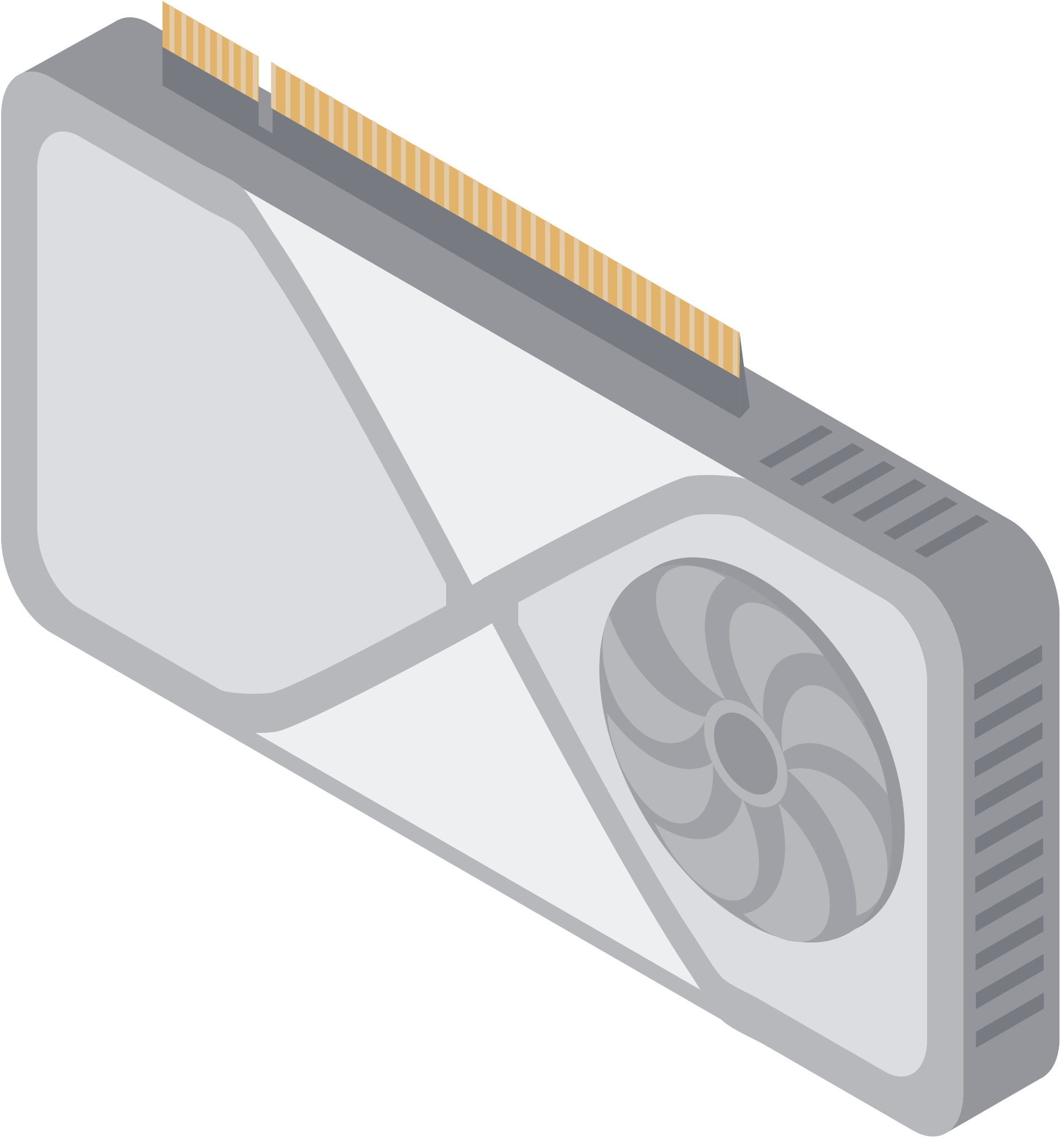}};
    \node[draw=gray!70, thick, rounded corners=2pt, inner sep=1pt,
          fit=(s-0-0)(s-3-31)] (sonicbox) {};
    \node[above=1mm of sonicbox, font=\large\bfseries, inner sep=0pt] {$>1000\times$ fewer GPU-h};
    \node[font=\footnotesize, inner sep=3pt, fill=white, fill opacity=0.8,
          rounded corners=2pt] at (sonicbox.center)
      {128 L40 GPUs, \qty{168}{\hour}~\cite{luo2025sonic}};
    \matrix[below=2pt of sonicbox.south east, anchor=north east, inner sep=0pt, column sep=.3333em] {%
      \node[anchor=center] (sm) {\includegraphics[width=\cellsz]{gpu_icon.pdf}}; &
      \node[anchor=center, font=\footnotesize, retcol] {1 H100 GPU, \qty{3}{\hour}}; \\
    };
    \draw[-{Latex[length=2.5mm]}, thick, gray!70, rounded corners=2pt]
      (s-0-0 |- sonicbox.south) |- (sm.west);
  \end{tikzpicture}
\end{minipage}
\caption{\textbf{A learned differentiable world model trains a humanoid whole-body controller and reduces GPU-hours by more than $1000\times$ versus cluster-scale model-free reinforcement learning (\hyperref[par:accounting]{\S\ref*{sec:scaling}}).} Left: the policy improves through imagined rollouts of the world model. Right: 128 L40 GPUs train for \qty{168}{\hour} in SONIC~\citep{luo2025sonic}, against one H100 for \qty{3}{\hour} in this work.}
\end{figure}

\begin{abstract}
Tracking large human-motion datasets has become the dominant route to general, expressive humanoid whole-body control, but scaling across many motions is expensive. Current pipelines use model-free reinforcement learning, either merging per-motion specialists through lossy distillation, or brute-forcing a single policy with cluster-scale compute.
SuperMimic amortizes learning across all motions with one differentiable world model of their shared dynamics, fit from logged transitions of a standard rigid-body simulator. The world model and the tracking policy train together in one loop, and backpropagation-through-time over short imagined rollouts gives the policy analytic gradients. Fast imagined rollouts and analytic gradients make policy learning efficient.
In 3 hours on a single GPU, SuperMimic surpasses the return of a state-of-the-art motion tracker baseline trained for 7 days on 128 GPUs. It reduces GPU-hours by more than $1000\times$ and fails on only 25 of \num{129514} benchmark motions, versus 179 failures for the baseline.  After 9.8 hours on two GPUs, SuperMimic achieves zero failures.
From 1 to 264 hours of motion, the world model's prediction error stays stable, and GPU-hours to a target return grow slowly. Our policy transfers to a physical Unitree~G1 across diverse and acrobatic motions.
SuperMimic removes compute as the bottleneck to scaling humanoid whole-body control, making it practical to measure motion-tracking scaling laws.
\end{abstract}


\section{Introduction}

Humanoid robots hold the promise of operating in environments designed for people, but doing so requires whole-body controllers that span diverse human motion, from locomotion and dance to crouching and balance recovery. Physics-based motion imitation addresses this by training RL policies to track reference motion clips \cite{peng2018deepmimic} drawn from large motion-capture (mocap) datasets \cite{luo2025sonic}. Existing systems scale this idea either by distilling many clip-specific experts into one controller \cite{liao2025beyondmimic} or by training a generalist policy over a large motion library \cite{luo2025sonic}. In both cases, a more diverse motion library produces a more capable controller. But both also become computationally expensive as motion coverage grows: the state-of-the-art tracker SONIC~\citep{luo2025sonic}, for example, trains for \qty{\sim 168}{\hour} on 128 GPUs, totaling roughly \num{21000} GPU-hours. Although reference motions differ, they share the same humanoid hybrid rigid-body dynamics. Model-free RL fails to explicitly model this shared structure, instead repeatedly estimating action consequences from noisy simulator rollouts. Scaling to more motions therefore amounts to scaling up compute.

SuperMimic addresses this inefficiency with a single learned, differentiable world model fit from logged simulator transitions. Backpropagation-through-time (BPTT) through short imagined rollouts provides analytic policy gradients through the learned model, allowing one dynamics model to amortize learning across the entire mocap library. Because these gradients come from the learned model rather than from differentiating the simulator, SuperMimic remains compatible with rigid-body simulators, including those with non-smooth contacts and non-differentiable implementations.

On one GPU, SuperMimic surpasses the normalized return of SONIC's released checkpoint within \qty{3}{\hour} and reduces GPU-hours by more than $1000\times$ (\hyperref[par:accounting]{\S\ref*{sec:scaling}}). On a \num{129514}-motion benchmark, SuperMimic fails on 25 motions after \qty{3}{\hour} on one GPU and on none after \qty{9.8}{\hour} on two GPUs, against 179 for SONIC's released checkpoint (Appendix~\ref{app:eval}). Across a $264\times$ increase in dataset length, the world model's prediction error stays stable and SuperMimic's compute to a fixed return grows slowly. The learned policy transfers to a physical Unitree~G1. In a preliminary study of downstream interfaces, an RL-trained velocity encoder outperforms a distillation baseline.

\paragraph{Contributions.}
\begin{itemize}[leftmargin=*,noitemsep,topsep=2pt]
  \item \textbf{Model-based RL (MBRL) for deployable humanoid control}. We demonstrate that BPTT through a learned, differentiable world model can train whole-body tracking policies in a setting long dominated by model-free proximal policy optimization (PPO)~\citep{schulman2017proximal}, and that the resulting controller transfers to a physical Unitree~G1.
  \item \textbf{Compute is no longer the bottleneck to scaling.} SuperMimic surpasses the return of SONIC, a cluster-scale PPO pipeline, on a single GPU and reduces GPU-hours by more than $1000\times$ (\hyperref[par:accounting]{\S\ref*{sec:scaling}}). Its training cost grows slowly as the motion library expands, making scaling humanoid whole-body control practical without huge GPU infrastructure.
  \item \textbf{An analysis of the ingredients behind the gains.} We identify three world-model properties behind the compute reduction: prediction error that stays stable over a $264\times$ increase in dataset length, rollouts more than $10\times$ faster than the simulator, and analytic gradients, with which the policy reaches the best return of PPO on the world model with about $64\times$ fewer imagined steps.
  \item \textbf{RL-trained downstream control beyond distillation.} As a preliminary study, the same learned world model trains a downstream goal encoder on the interface's own task reward. On velocity commands, the policy tracks more accurately with this encoder than with a distillation baseline.
\end{itemize}

\section{Related Work}

\begin{table}[t]
\centering\small
\setlength{\tabcolsep}{6pt}
\setlength{\aboverulesep}{0pt}
\setlength{\belowrulesep}{0pt}
\setlength{\extrarowheight}{0.4ex}
\begin{tabular}{@{}lccccc@{}}
\toprule
Method & Mimic & Latent & Downstream & Task signal & Real robot \\
\midrule
NPMP~\citep{merel2018neural}                & SVG(0) & BC     & SVG(0)      & RL                & \xmarkr \\
ControlVAE~\citep{yao2022controlvae}        & \multicolumn{2}{>{\columncolor{AnaFill}}c}{MBRL}     & MBRL    & RL                & \xmarkr \\
PhysicsVAE~\citep{won2022physics}           & PPO   & BC$^1$ & PPO         & RL                & \xmarkr \\
PULSE~\citep{luo2024universal}              & PPO   & DAgger & PPO         & RL                & \xmarkr \\
MaskedMimic~\citep{tessler2024maskedmimic}  & A2C   & \multicolumn{2}{>{\columncolor{AnaFill}}c}{DAgger} & Distill     & \xmarkr \\
BeyondMimic~\citep{liao2025beyondmimic}     & PPO   & DAgger & CGD         & Distill         & \cmarkg \\
SONIC~\citep{luo2025sonic}                  & \multicolumn{3}{>{\columncolor{AnaFill}}c}{PPO}      & Distill$^2$ & \cmarkg \\
LATENT~\citep{zhang2026learning}            & PPO   & DAgger & PPO         & RL                & \cmarkg \\
\midrule
\textbf{SuperMimic}          & \multicolumn{3}{>{\columncolor{AnaFill}}c}{MBRL}              & Distill\ /\ RL                & \cmarkg \\
\bottomrule
\end{tabular}
\caption{\textbf{Comparison of various methods and their training strategies across different stages.} The Mimic, Latent, and Downstream columns indicate the training strategy used at each stage. Task signal is what trains the downstream interface: its own task reward (RL) or a reference-tracking signal (Distill). Merged cells denote jointly trained stages, while adjacent unmerged cells indicate sequential training, with each stage trained after freezing the previous one.\protect\newline SVG(0): stochastic value gradients~\citep{heess2015learning}; BC: offline behavior cloning on fixed expert rollouts, as opposed to DAgger's student rollouts; CGD: classifier guidance diffusion; A2C: advantage actor--critic; DAgger: dataset aggregation. $^1$Behavior cloning with a one-step next-state consistency term through a frozen, learned dynamics model. $^2$Goal encoders train jointly with the tracker on the motion-tracking reward and latent-alignment losses; no task-specific reward.}
\label{tab:related}
\end{table}

\textbf{Large-scale physics-based motion tracking.}
Physics-based motion imitation learns controllers that reproduce reference motions under contact-rich rigid-body dynamics.
DeepMimic~\citep{peng2018deepmimic} demonstrated expressive character skills through model-free reinforcement learning, and PHC~\citep{luo2023perpetual} extended this paradigm to large motion sets with robust recovery.
SONIC~\citep{luo2025sonic} further demonstrated that scaling motion data, policy capacity, and training compute produces generalist controllers that transfer to humanoid hardware.
These results establish motion tracking as a practical route to broad whole-body behavior coverage.
However, training these large-scale motion trackers requires extensive simulator rollouts and substantial computation.
SuperMimic addresses this bottleneck by learning a differentiable dynamics model shared across all motions, so policy improvement can exploit analytic gradients rather than model-free rollout estimates.

\textbf{Learned differentiable dynamics for motion control.}
SuperTrack~\citep{fussell2021supertrack} optimizes motion-control policies through a learned differentiable surrogate of the simulator.
ControlVAE~\citep{yao2022controlvae} jointly learns a dynamics model, latent skill representation, and generative controller, while MoConVQ~\citep{yao2024moconvq} extends learned-model training to discrete motion representations over tens of hours of data.
SuperDyno~\citep{lei2025scalable} scales differentiable neural dynamics to tens of thousands of motion clips.
These methods establish the effectiveness of learned dynamics and analytic policy gradients for scalable simulated-character control.
On hardware, a quadruped controller trained through a learned world model imitates animal motions~\citep{shi2024efficient}.
SuperMimic extends this direction to single-GPU training of a large-scale humanoid tracker and to its sim-to-real deployment: its world model conditions on simulator context, while the deployed actor uses noisy proprioceptive observations and a goal input (\S\ref{sec:wm}).
We evaluate how model accuracy, rollout throughput, and differentiable policy updates contribute to training efficiency, and how prediction error behaves as motion coverage grows (\S\ref{sec:scaling}--\ref{sec:wm-properties}), and demonstrate physical deployment of the resulting controller (\S\ref{sec:real-robot}).

\textbf{Model-based policy optimization and robot deployment.}
Dreamer~\citep{hafner2019dream} and DreamerV3~\citep{hafner2025dreamerv3} learn latent dynamics for imagination-based actor--critic learning, while TD-MPC~\citep{hansen2022temporal} combines short-horizon planning with terminal-value prediction.
DayDreamer~\citep{wu2023daydreamer} demonstrates world-model learning directly on physical robots.
SHAC~\citep{xu2022accelerated} uses differentiable simulation for short-horizon actor--critic updates, motivating SuperMimic's critic-bootstrapped objective.
Analytic-gradient robot control can also combine a simplified differentiable surrogate with a high-fidelity forward simulator, as demonstrated for quadruped locomotion~\citep{song2025differentiable}.
DMO~\citep{amigo2025first} takes a decoupled approach: it rolls out trajectories in the simulator but backpropagates through a learned dynamics model.
This mitigates compounding model error, and DMO reports an order-of-magnitude sample-efficiency gain over PPO.
However, every policy update still needs fresh simulator rollouts, and each policy learns a single task.
SuperMimic instead both rolls out and backpropagates through one learned world model shared across many motions.
It accepts some model error in exchange for rollouts that do not depend on simulator throughput (\S\ref{sec:wm-properties}) and are amortized across motions.

RWM~\citep{li2025robotic} and RWM-U~\citep{li2025uncertainty} demonstrate that policies trained in learned neural simulators can transfer to physical robots.
Both optimize policies using PPO on imagined rollouts; RWM-U additionally uses uncertainty-penalized offline optimization.
SuperMimic instead optimizes the policy by differentiating the return through its learned dynamics, as do MAAC~\citep{clavera2020model} and PWM~\citep{georgiev2025pwm}.
PWM, RWM, and RWM-U pretrain the world model on trajectories from other policies. In contrast, SuperMimic fits its world model during training on data its own policy collects.

\textbf{Reusable motion representations and downstream control.}
Reusable controllers connect motion tracking to downstream goal interfaces (Table~\ref{tab:related}).
NPMP~\citep{merel2018neural} and PhysicsVAE~\citep{won2022physics} learn latent motor representations from expert behavior; PULSE~\citep{luo2024universal} and LATENT~\citep{zhang2026learning} use learned motion representations for downstream reinforcement learning.
MaskedMimic~\citep{tessler2024maskedmimic} supports partial-body and multimodal goals through motion inpainting, while BeyondMimic~\citep{liao2025beyondmimic} uses guided diffusion for goal following.
SONIC~\citep{luo2025sonic} jointly trains goal encoders with its tracker using tracking rewards and latent-alignment losses.
ControlVAE~\citep{yao2022controlvae} optimizes downstream task policies through its learned model while freezing the skill controller.
SuperMimic jointly trains goal-specific encoders and a shared decoder through the same dynamics model (Appendix~\ref{app:unified}).
Goal interfaces can optimize either a reference-tracking reward or their own task reward.
Our preliminary study trains a velocity interface concurrently with the motion-tracking interface, once on each objective (\S\ref{sec:downstream}).

\section{Method}
\label{sec:method}

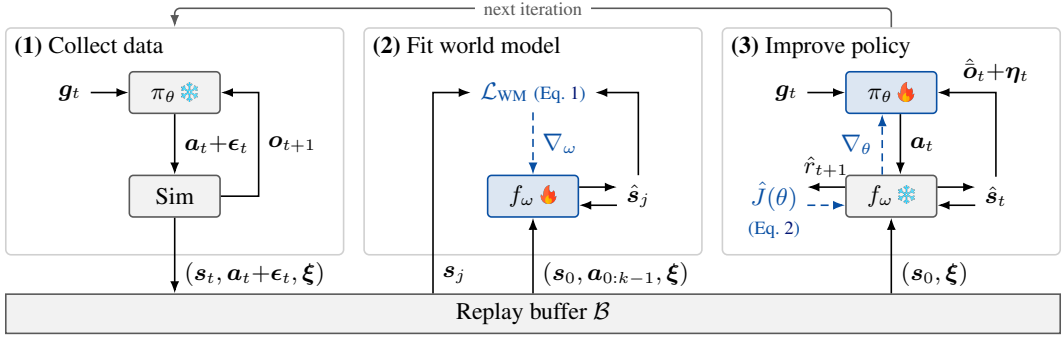
\begin{figure}[t]
\centering
\def\panelgap{2.5mm}%
\def\gutter{5mm}%
\def\lane{1.2mm}%
\def\iconh{2.4mm}%
\def\eqtag#1{{\scriptsize(Eq.~\ref{#1})}}%
\begin{tikzpicture}[
  font=\footnotesize, semithick, >={Latex[length=1.6mm]},
  node distance=8mm and 5mm,
  module/.style={draw=context, fill=region, rounded corners=1.5pt, minimum width=12mm, minimum height=5.5mm},
  trained/.style={draw=ours, fill=highlight},
  buffer/.style={draw=context, fill=region, minimum width=\textwidth-\pgflinewidth, minimum height=5mm},
  panel/.style={draw=track, rounded corners=2pt, minimum width=(\textwidth-\panelgap*2)/3-\pgflinewidth, minimum height=30mm, label={[anchor=north west]north west:#1}},
  grad/.style={->, ours, densely dashed},
  lane/.style={transform canvas={shift={(#1:\lane)}}},
]
\node[panel={\textbf{(1)} Collect data}] (P1) {};
\node[panel={\textbf{(2)} Fit world model}, right=\panelgap of P1] (P2) {};
\node[panel={\textbf{(3)} Improve policy}, right=\panelgap of P2] (P3) {};
\node[buffer, below=\gutter of P2] (buf) {Replay buffer $\mathcal{B}$};

\node[module, above=\gutter of P1.south] (sim) {Sim};
\node[module, above=of sim] (pi1) {$\pi_\theta$\,\frozenicon[\iconh]};
\node[left=of pi1] (g1) {$\bm{g}_t$};
\coordinate[right=of sim] (o1);
\draw[->] (g1) -- (pi1);
\draw[->] (pi1) -- node[right]{$\bm{a}_t{+}\bm{\epsilon}_t$} (sim);
\draw[->] (sim) -- (o1) -- node[right]{$\bm{o}_{t+1}$} (o1 |- pi1) -- (pi1);
\draw[->] (sim) -- (sim |- P1.south) -- node[right]{$(\bm{s}_t,\bm{a}_t{+}\bm{\epsilon}_t,\bm{\xi})$} (sim |- buf.north);

\node[module, trained, above=\gutter of P2.south] (wm2) {$f_\omega$\,\trainicon[\iconh]};
\node[ours] (loss) at (wm2 |- pi1) {$\mathcal{L}_{\text{WM}}$ \eqtag{eq:wm-loss}};
\node[right=of wm2] (sh2) {$\hat{\bm{s}}_j$};
\coordinate[left=of loss] (in2);
\draw[->] (in2 |- buf.north) -- node[right]{$\bm{s}_j$} (in2 |- P2.south) -- (in2) -- (loss);
\draw[->] (sh2) -- (sh2 |- loss) -- (loss);
\draw[grad] (loss) -- node[right]{$\nabla_\omega$} (wm2);
\draw[->, lane=90] (wm2) -- (sh2);
\draw[->, lane=-90] (sh2) -- (wm2);
\draw[->] (wm2 |- buf.north) -- node[right]{$(\bm{s}_0,\bm{a}_{0:k-1},\bm{\xi})$} (wm2 |- P2.south) -- (wm2);

\node[module, above=\gutter of P3.south] (wm3) {$f_\omega$\,\frozenicon[\iconh]};
\node[module, trained, above=of wm3] (pi3) {$\pi_\theta$\,\trainicon[\iconh]};
\node[left=of pi3] (g3) {$\bm{g}_t$};
\node[ours, left=of wm3, label={[ours, inner sep=0pt]below:\eqtag{eq:policy-obj}}] (J) {$\hat{J}(\theta)$};
\node[right=of wm3] (sh3) {$\hat{\bm{s}}_t$};
\draw[->] (g3) -- (pi3);
\draw[->, lane=0] (pi3) -- node[right]{$\bm{a}_t$} (wm3);
\draw[->] (sh3) -- (sh3 |- pi3) -- node[above, pos=0.0]{$\hat{\bar{\bm{o}}}_t{+}\bm{\eta}_t$} (pi3);
\draw[->, lane=90] (wm3) -- (sh3);
\draw[->, lane=-90] (sh3) -- (wm3);
\draw[->, lane=90] (wm3) -- node[above]{$\hat{r}_{t+1}$} (J);
\draw[grad, lane=-90] (J) -- (wm3);
\draw[grad, lane=180] (wm3) -- node[left]{$\nabla_\theta$} (pi3);
\draw[->] (wm3 |- buf.north) -- node[right]{$(\bm{s}_0,\bm{\xi})$} (wm3 |- P3.south) -- (wm3);

\coordinate[above=\panelgap of P3.north] (next);
\draw[->, context, rounded corners=2pt] (P3) -- (next) -- node[fill=white, font=\scriptsize]{next iteration} (next -| P1) -- (P1);
\end{tikzpicture}%
\caption{\textbf{SuperMimic training loop.} (1)~The current policy $\pi_\theta$ collects transitions in simulation into a fixed-size replay buffer $\mathcal{B}$, where new transitions replace the oldest. (2)~The differentiable world model is fit to $k$-step windows sampled from $\mathcal{B}$ (\S\ref{sec:wm}). (3)~The policy is improved by backpropagation-through-time through the frozen world model over $H$ imagined steps; a critic $V_\psi$, trained in the same phase, bootstraps the return beyond $H$ (\S\ref{sec:policy}). Dashed arrows are gradients; \frozenicon[0.28cm]~=~frozen, \trainicon[0.28cm]~=~trained. Symbols: goal $\bm{g}_t$ (e.g., reference motion), action $\bm{a}_t$, action noise $\bm{\epsilon}_t$, observation $\bm{o}_t$, state $\bm{s}$ and predicted state $\hat{\bm{s}}$, and sim context $\bm{\xi}$ (domain-randomization parameters). $\mathcal{L}_{\text{WM}}$ is the world-model loss, $\hat{\bar{\bm{o}}}_t$ the predicted noise-free observation, $\bm{\eta}_t$ the sensor noise, $\hat{r}_{t+1}$ the predicted reward, and $\hat{J}(\theta)$ the policy objective.}
\label{fig:pipeline}
\end{figure}

\subsection{SuperMimic training loop}
\label{sec:problem}
SuperMimic trains one goal-conditioned whole-body controller across a multi-skill motion library and supports multiple goal interfaces. It captures the shared humanoid dynamics in a differentiable world model and improves the policy through short imagined rollouts. Training alternates between three phases in one loop (Figure~\ref{fig:pipeline}). First, the current policy collects simulator transitions into a replay buffer. Second, we fit the differentiable world model to short replay windows. Third, the policy is improved by backpropagating an imagined return through the frozen world model, with each rollout seeded from a replay-buffer state in the Dyna style~\citep{sutton1991dyna, janner2019trust}. The buffer, model, and policy are updated throughout training, so dynamics learning and control improve together rather than in separate stages (Algorithm~\ref{alg:train}, Appendix~\ref{app:algorithm}). Transitions from earlier policies remain valid for fitting the world model, because the dynamics do not depend on the policy that collected them.

We denote the goal-conditioned partially observable Markov decision process (POMDP) by $\mathcal{M}$ and use compact notation with state $\bm{s}_t$, goal $\bm{g}_t$, and action $\bm{a}_t$. State $\bm{s}_t$ gives the proprioceptive observation $\bm{o}_t$ and the termination $d_t$, which is 1 if $\bm{s}_t$ is terminal. Action $\bm{a}_t$ in state $\bm{s}_t$ yields the reward $r_{t+1}$ and the next state $\bm{s}_{t+1}$. The policy is deterministic, $\bm{a}_t=\pi_\theta(\bm{o}_t,\bm{g}_t)$, and $\bm{a}_t$ parameterizes a normalized joint setpoint; for motion tracking, $\bm{g}_t$ is read from a reference motion. During data collection, the simulator executes the noisy action $\bm{a}_t+\bm{\epsilon}_t$ with action noise $\bm{\epsilon}_t\sim\rho_\epsilon$, and the replay buffer stores this executed noisy action. The noise widens the range of actions on which the world model is fit; it serves only the model fit and is optional. Returns use a soft survival weight, $c_0=1$ and $c_{t+1}=c_t(1-d_{t+1})$, giving $J(\theta)=\mathbb{E}[\sum_{t\ge0}\gamma^t c_t r_{t+1}]$ with discount $\gamma\in[0,1)$. The simulator logs hard terminations, while the learned model predicts a termination probability $\hat{d}_t\in[0,1]$, which keeps the survival weight differentiable in imagined rollouts. Throughout, a hat marks the world-model counterpart of a simulator quantity. We also factor the observation as $\bm{o}_t=\bar{\bm{o}}_t+\bm{\eta}_t$, with $\bar{\bm{o}}_t$ the noise-free observation and known observation noise $\bm{\eta}_t\sim\rho_\eta$ sampled outside the world model. Appendix~\ref{app:spaces} gives the full spaces and reward.

The default interface is full-reference motion tracking. Other goal interfaces, such as velocity, share the same learned dynamics and train jointly with motion tracking (Appendix~\ref{app:unified}).

\subsection{Learning a differentiable world model}
\label{sec:wm}
The simulator's articulated multi-rigid-body dynamics are known, but the simulator need not provide gradients, so SuperMimic learns a differentiable surrogate to obtain policy gradients: it supplies the predictions and Jacobians for the BPTT update in \S\ref{sec:policy}. The resulting analytic policy gradient is the gradient of a surrogate for the expected return in simulation. Since the simulator's randomness has known sources, the world model stays deterministic: it conditions on the per-episode sim context $\bm{\xi}$ (friction, calibration offsets, and torso center-of-mass) and predicts the noise-free observation $\bar{\bm{o}}_t$.

The world model contains a learned dynamics model and reward, observation, and termination heads; each head is either learned or computed analytically,
\[
  \hat{\bm{s}}_{t+1}=f_\omega(\hat{\bm{s}}_t,\bm{a}_t,\bm{\xi}),\qquad
  \hat{r}_{t+1}=r_\omega(\hat{\bm{s}}_t,\bm{a}_t,\bm{g}_t),\qquad
  \hat{\bar{\bm{o}}}_t=g_\omega(\hat{\bm{s}}_t,\bm{\xi}),\qquad
  \hat{d}_t=d_\omega(\hat{\bm{s}}_t,\bm{g}_t).
\]
We fit the network weights $\omega$ on $k$-step autoregressive replay windows ($k{=}8$), seeded at the logged state $\hat{\bm{s}}_0=\bm{s}_0$ and rolled forward on logged actions, by minimizing
\begin{equation}
\label{eq:wm-loss}
  \mathcal{L}_{\text{WM}}(\omega) = \mathbb{E}_{\mathcal{B}} \sum_{j=1}^{k} \Big[ \lVert \hat{\bm{s}}_{j} - \bm{s}_{j} \rVert^2 + (\hat{r}_{j} - r_{j})^2 + \lVert \hat{\bar{\bm{o}}}_{j} - \bar{\bm{o}}_{j} \rVert^2 + \mathrm{BCE}(\hat{d}_{j}, d_{j}) \Big],
\end{equation}
where an analytic head has no loss term. Here $r_j$ is the simulator reward, $d_j\in\{0,1\}$ is the simulator termination label, and BCE is the binary cross-entropy.

\subsection{Learning the policy through the world model}
\label{sec:policy}
During each policy update, we hold the world-model weights $\omega$ fixed and optimize $\pi_\theta$ and $V_\psi$ on imagined rollouts of $H{=}32$ steps. The critic $V_\psi$ estimates the return beyond step $H$, so the model must be accurate only within these $H$ steps. This imagination horizon exceeds the $k$-step replay windows on which we fit the model (\S\ref{sec:wm}). In practice, the learned model remains stable over indefinitely long rollouts, because the $k$-step autoregressive fit trains it to correct its own prediction errors.

Each imagined rollout starts from a replay-buffer state $\bm{s}_0\sim\rho_0$ with its recorded sim context $\bm{\xi}$. At step $t$, the model emits $\hat{\bar{\bm{o}}}_t$, we re-add sensor noise, and the policy acts:
$\bm{a}_t=\pi_\theta(\hat{\bar{\bm{o}}}_t+\bm{\eta}_t,\bm{g}_t)$, $\bm{\eta}_t\sim\rho_\eta$.
The policy objective over imagined rollouts is
\begin{equation}
\label{eq:policy-obj}
  \hat{J}(\theta) = \mathbb{E}_{\bm{s}_0 \sim \rho_0,\, \pi_\theta} \Big[ \sum_{t=0}^{H-1} \gamma^t \hat{c}_t\, \hat{r}_{t+1} ~~+~~ \gamma^H \hat{c}_H\, V_\psi(\hat{\bm{s}}_H, \bm{g}_H, \bm{\xi}) \Big].
\end{equation}
For motion tracking, the reward $r_{t+1}$ is the BeyondMimic motion-tracking reward~\citep{liao2025beyondmimic} without its self-collision penalty (Appendix~\ref{app:spaces}).

The world model is differentiable, so BPTT computes a first-order gradient $\nabla_\theta\hat{J}$ through the $H$ imagined steps via the world model's Jacobians with respect to $\hat{\bm{s}}_t$ and $\bm{a}_t$. In contrast, PPO treats the dynamics as a black box: it weights the score $\nabla_\theta\log\pi_\theta$ of a stochastic policy by the advantages of sampled actions, a likelihood-ratio estimate. BPTT instead differentiates the return with respect to each action, so over short horizons its gradient has much lower variance.

The critic $V_\psi$ is fit to generalized advantage estimation returns~\citep{schulman2015high} computed backward in time over the imagined transitions only,
\begin{equation}
\label{eq:critic}
  \mathcal{L}_V(\psi) = \mathbb{E} \sum_{t=0}^{H-1} \big( V_t - \hat{R}_t \big)^2, \qquad
  \hat{R}_t = \hat{r}_{t+1} + \gamma (1-\hat{d}_{t+1}) \big[ (1-\lambda) V_{t+1} + \lambda \hat{R}_{t+1} \big],
\end{equation}
with $V_t = V_\psi(\hat{\bm{s}}_t, \bm{g}_t, \bm{\xi})$. The recursion starts from $\hat{R}_H = V_H$, the critic's estimate of the return beyond the horizon. The targets $\hat{R}_t$ are computed once per rollout and remain fixed while $\psi$ is fit.

The policy and critic updates together form a short-horizon actor--critic in the spirit of SHAC~\citep{xu2022accelerated}, which backpropagates through a differentiable simulator. SuperMimic replaces SHAC's hard episode termination with the survival weight $\hat{c}_t$, and its critic has no exponential-moving-average target network. Prior motion-tracking controllers trained with analytic gradients often omit a critic and optimize only the $H$-step reward sum~\citep{fussell2021supertrack, yao2022controlvae, ren2023diffmimic}. Without a critic, training can reach a similar nominal tracking return, but this unbootstrapped objective is more myopic. The critic matters after perturbations, where recovery or balance rewards can lie beyond the imagination window (\S\ref{sec:critic}).

\section{Experimental Results}

We show that SuperMimic scales favorably relative to PPO pipelines, identify the world-model properties behind this scaling, demonstrate transfer on a real Unitree~G1, and report a preliminary study of downstream goal interfaces. The simulator is MuJoCo Warp~\citep{mujocowarp2025}, a GPU implementation of MuJoCo~\citep{todorov2012mujoco}. All our reported costs are measured on NVIDIA~H100 (\qty{80}{\giga\byte} SXM5) GPUs of Lambda on-demand instances. We ran several random seeds and found training very stable, which we attribute to the dense reward that motion tracking gives at every step. The figures therefore do not report seed variation. Normalized return, return for short, is the undiscounted sum of the per-step reward over the full episode, divided by the largest value this sum can take. A policy that terminates early therefore scores low.

\subsection{SuperMimic reaches SONIC-level tracking with far fewer GPU-hours}
\label{sec:scaling}

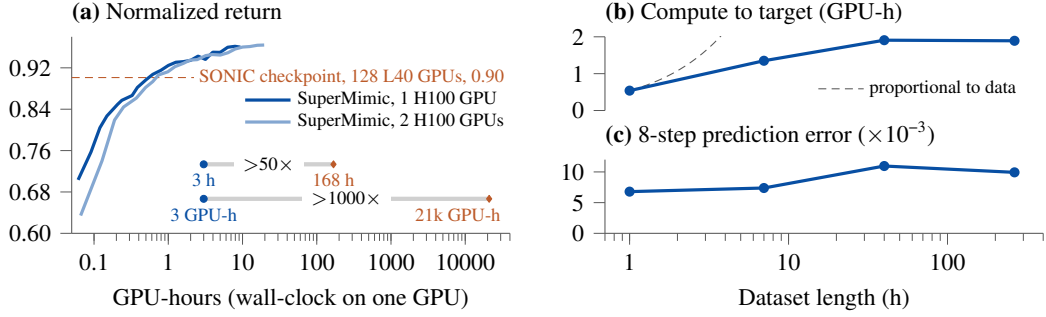
\begin{figure}[t]
\centering
\def\budgetrows{10.5mm}%
\begin{tikzpicture}[font=\scriptsize]
\begin{axis}[
    name=curve, smplot, width=\plotequal, height=\plottall,
    title={\textbf{(a)} Normalized return},
    xlabel={GPU-hours (wall-clock on one GPU)},
    xmode=log, xmin=0.05, xmax=4e4,
    xtick={0.1,1,10,100,1000,10000,100000}, xticklabels={0.1,1,10,100,1000,10000,},
    ymin=0.6, ymax=0.98, ytick={0.6,0.68,0.76,0.84,0.92}, yticklabel style={/pgf/number format/fixed, /pgf/number format/fixed zerofill, /pgf/number format/precision=2},
    legend style={at={(axis cs:4e4,0.88)}, anchor=north east, inner sep=0pt, nodes={inner ysep=1pt}},
]
\addplot[theirs, densely dashed, forget plot] coordinates {(0.05,0.9012) (4e4,0.9012)} node[pos=1, left, fill=white, inner sep=2pt] {SONIC checkpoint, 128 L40 GPUs, 0.90};
\addplot[ours, very thick] coordinates {
      (0.061,0.7034) (0.091,0.7573) (0.120,0.8042) (0.149,0.8269) (0.208,0.8495) (0.237,0.8572) (0.326,0.8665) (0.384,0.8821)
      (0.502,0.8970) (0.619,0.9074) (0.795,0.9152) (1.001,0.9246) (1.264,0.9308) (1.586,0.9319) (1.995,0.9344) (2.521,0.9426)
      (3.164,0.9363) (3.982,0.9502) (5.006,0.9499) (6.322,0.9598) (7.932,0.9616) (9.368,0.9600)
    };
\addlegendentry{SuperMimic, 1 H100 GPU}
\addplot[ours!50, very thick] coordinates {
      (0.066,0.6344) (0.128,0.7401) (0.189,0.8191) (0.250,0.8443) (0.372,0.8616) (0.495,0.8821) (0.617,0.8933) (0.740,0.9063)
      (0.985,0.9123) (1.230,0.9223) (1.598,0.9287) (1.965,0.9349) (2.454,0.9359) (3.188,0.9436) (3.921,0.9457) (5.022,0.9447)
      (6.367,0.9492) (7.956,0.9564) (10.036,0.9602) (12.606,0.9612) (15.788,0.9636) (19.584,0.9641)
    };
\addlegendentry{SuperMimic, 2 H100 GPUs}
\end{axis}
\begin{axis}[
    smplot, hide axis, width=\plotequal, height=\budgetrows,
    at={(curve.south west)}, anchor=south west,
    xmode=log, xmin=0.05, xmax=4e4,
    ymin=0, ymax=2.3,
]
\addplot[track, ultra thick] coordinates {(3,2) (168,2)} node[midway, fill=white, text=black] {$>$50$\times$};
\addplot[track, ultra thick] coordinates {(3,1) (21000,1)} node[midway, fill=white, text=black] {$>$1000$\times$};
\addplot[ours, only marks, mark=*] coordinates {(3,2) (3,1)};
\addplot[theirs, only marks, mark=diamond*] coordinates {(168,2) (21000,1)};
\node[ours, below] at (axis cs:3,2) {3 h};
\node[theirs, below] at (axis cs:168,2) {168 h};
\node[ours, below] at (axis cs:3,1) {3 GPU-h};
\node[theirs, below left] at (axis cs:4e4,1) {21k GPU-h};
\end{axis}
\begin{axis}[
    name=cost, smplot, width=\plotequal, height=\plothalf,
    at={(curve.north east)}, anchor=north west, xshift=\plotgap,
    title={\textbf{(b)} Compute to target (GPU-h)},
    xmode=log, xmin=0.7, xmax=400, axis x line=none,
    ymin=0, ymax=2, ytick={0,1,2},
    legend pos=south east,
]
\addplot[context, densely dashed, domain=1:7] {0.539*x};
\addlegendentry{proportional to data}
\addplot[ours, very thick, mark=*] coordinates {(1,0.539) (7,1.350) (40,1.908) (264,1.892)};
\end{axis}
\begin{axis}[
    smplot, width=\plotequal, height=\plothalf,
    at={(cost.south west)}, anchor=north west, yshift=-\plotsep,
    title={\textbf{(c)} 8-step prediction error ($\times$10\textsuperscript{\textminus3})},
    xlabel={Dataset length (h)},
    xmode=log, xmin=0.7, xmax=400, xtick={0.1,1,10,100,1000}, xticklabels={,1,10,100,},
    ymin=0, ymax=0.012, ytick={0,0.005,0.01}, scaled y ticks=base 10:3, ytick scale label code/.code={},
]
\addplot[ours, very thick, mark=*] coordinates {(1,0.00679) (7,0.00738) (40,0.01096) (264,0.00993)};
\end{axis}
\end{tikzpicture}
\caption{\textbf{SuperMimic surpasses SONIC's tracking return with far fewer GPU-hours.} (a)~SuperMimic on one and two H100 GPUs surpasses the return of the released SONIC checkpoint (dashed), all three scored on the same benchmark of \num{129514} motions (\qty{264}{\hour}). Two budget rows compare SuperMimic (dot) with SONIC (diamond): SuperMimic reduces wall-clock time by more than $50\times$ and GPU-hours by more than $1000\times$. (b, c)~On a shared dataset-length sweep, GPU-hours to target return (b) grow slowly while the world model's 8-step prediction error (c) stays within 0.007--0.011 as the dataset grows from 1 to \qty{264}{\hour}, consistent with one world model amortizing shared body and contact dynamics across motions. The prediction error counts the state only, is normalized by a predefined nominal error, and is measured online on policy rollouts.}
\label{fig:compute}
\end{figure}

We compare against SONIC, a published PPO pipeline trained on a 128-GPU cluster. Both methods build on the BeyondMimic motion-tracking formulation~\citep{liao2025beyondmimic}, making SONIC the closest available published comparison. Within \qty{3}{\hour} on one GPU, SuperMimic surpasses the normalized return of the released SONIC checkpoint (Figure~\ref{fig:compute}(a)). On a shared benchmark of \num{129514} motions, it also fails on fewer motions than SONIC: 25 after \qty{3}{\hour} on one GPU and 0 after \qty{9.8}{\hour} on two GPUs, against 179 for SONIC's released checkpoint (Appendix~\ref{app:eval}). We state the \qty{3}{\hour} conservatively: from then on, every checkpoint is ahead of SONIC by at least half of the gap between SONIC and SuperMimic's best result, in both return and failed motions. SuperMimic thus uses more than $50\times$ less wall-clock time and more than $1000\times$ fewer GPU-hours than SONIC.

\phantomsection\label{par:accounting}\noindent\textbf{Compute accounting.} The ratios compare SONIC's reported 128 NVIDIA~L40 GPUs for 7 days (about \num{21000} GPU-hours) with \qty{3}{\hour} on one NVIDIA~H100, within which SuperMimic surpasses the return of SONIC's released checkpoint. They are about $56\times$ ($>50\times$) in wall-clock time and about $7200\times$ ($>1000\times$) in GPU-hours. The \qty{3}{\hour} cover the full training loop; checkpoint evaluations run after training and are not counted. SuperMimic trains on \qty{264.56}{\hour} of motion from BONES-SEED~\citep{bones2026seed}, the public subset of SONIC's data, about $2.3\times$ less than SONIC's \qty{611}{\hour}. The GPU models also differ: in peak throughput, an H100 delivers $0.74\times$ the fp32 and $5.5\times$ the tensor-core performance of an L40; how much of that peak either run used is unknown. Nonetheless, $2.3\times$ in data and at most $5.5\times$ in peak GPU throughput cannot account for a $1000\times$ gap in GPU-hours.

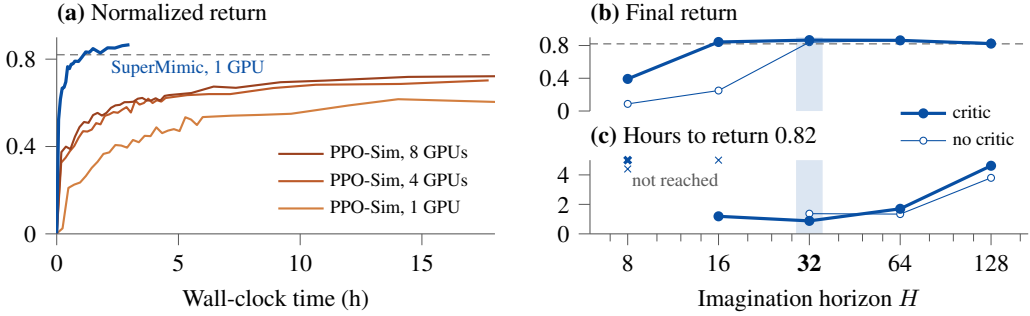
\begin{figure}[t]
\centering
\begin{tikzpicture}[font=\scriptsize]
\begin{axis}[
    name=race, smplot, width=\plotequal, height=\plottall,
    title={\textbf{(a)} Normalized return},
    xlabel={Wall-clock time (h)},
    xmin=0, xmax=18,
    return axis,
    legend pos=south east, reverse legend,
]
\addplot[context, densely dashed, forget plot] coordinates {(0,0.82) (18,0.82)};
\addplot[theirslight, thick] coordinates {(0.00236111,0.001764) (0.24,0.025203) (0.479583,0.21044) (0.720972,0.2256) (0.960417,0.2352) (1.19861,0.2651) (1.43987,0.3023) (1.67875,0.3301) (1.91973,0.3662) (2.16125,0.3731) (2.40778,0.4068) (2.65042,0.4044) (2.88959,0.3948) (3.1318,0.43) (3.37181,0.4195) (3.61264,0.449) (3.85361,0.4474) (4.09305,0.4878) (4.32972,0.4612) (4.57042,0.4728) (4.81069,0.4795) (5.0457,0.4707) (5.28334,0.534) (5.52069,0.5233) (5.75889,0.4984) (5.99695,0.5348) (7.18611,0.5412) (8.38444,0.5449) (9.62625,0.5499) (11.9934,0.5881) (14.0222,0.6166) (18.6919,0.602)};
\addlegendentry{PPO-Sim, 1 GPU}
\addplot[theirs, thick] coordinates {(0.00180555,0.00181) (0.184722,0.3262) (0.365278,0.3488) (0.542777,0.3788) (0.721389,0.4025) (0.900417,0.4404) (1.08069,0.4702) (1.25958,0.4685) (1.43903,0.4776) (1.61959,0.5071) (1.7975,0.5015) (1.97611,0.5406) (2.15487,0.5471) (2.33319,0.5577) (2.51125,0.5544) (2.68931,0.5685) (2.86806,0.5803) (3.04625,0.5556) (3.22445,0.6101) (3.40278,0.5919) (3.58125,0.5994) (3.75931,0.619) (3.93736,0.599) (4.11569,0.6122) (4.29402,0.6056) (4.47208,0.6221) (5.35902,0.6353) (6.24459,0.64) (7.12736,0.64) (8.89098,0.6677) (10.646,0.6828) (14.1303,0.6865) (17.7414,0.7028)};
\addlegendentry{PPO-Sim, 4 GPUs}
\addplot[theirsdark, thick] coordinates {(0.00138889,0.001772) (0.19125,0.3744) (0.377223,0.3994) (0.563195,0.388) (0.748611,0.4448) (0.934584,0.4885) (1.12069,0.5124) (1.30653,0.5082) (1.49194,0.5438) (1.67639,0.554) (1.86055,0.5414) (2.04486,0.5543) (2.22903,0.5772) (2.41305,0.5883) (2.59667,0.5909) (2.78097,0.603) (2.96445,0.6032) (3.14792,0.605) (3.33125,0.6189) (3.51472,0.6225) (3.69819,0.6107) (3.88138,0.6108) (4.06458,0.621) (4.24833,0.6094) (4.43153,0.6318) (4.61486,0.6354) (5.53014,0.6449) (6.445,0.6744) (7.36055,0.6688) (9.19111,0.6944) (11.0203,0.7018) (14.4397,0.7188) (18.0328,0.7218)};
\addlegendentry{PPO-Sim, 8 GPUs}
\addplot[ours, very thick, forget plot] coordinates {(0.0021125,0.001519) (0.077808,0.524) (0.151899,0.6166) (0.225991,0.6672) (0.299996,0.67) (0.373919,0.696) (0.44801,0.7656) (0.522101,0.7565) (0.596276,0.777) (0.670791,0.779) (0.745049,0.7909) (0.894413,0.7894) (1.04353,0.8102) (1.19272,0.8331) (1.342,0.8309) (1.49103,0.8483) (1.79009,0.8282) (2.08984,0.8519) (2.38983,0.8509) (2.69008,0.8616) (2.98771,0.8661)};
\node[ours, below right, inner sep=2pt] at (axis cs:2,0.82) {SuperMimic, 1 GPU};
\end{axis}
\begin{axis}[
    name=return, smplot, width=\plotequal, height=\plothalf,
    at={(race.north east)}, anchor=north west, xshift=\plotgap,
    title={\textbf{(b)} Final return},
    xmode=log, xmin=6, xmax=170, axis x line=none,
    return axis,
]
\draw[highlight, line width=1em] (axis cs:32,0) -- (axis cs:32,0.9);
\addplot[context, densely dashed] coordinates {(6,0.82) (170,0.82)};
\addplot[ours, mark=*, mark options={fill=white}] coordinates {(8,0.087) (16,0.25) (32,0.84) (64,0.862) (128,0.835)};
\addplot[ours, very thick, mark=*] coordinates {(8,0.392) (16,0.843) (32,0.866) (64,0.863) (128,0.823)};
\end{axis}
\begin{axis}[
    smplot, axis on top, width=\plotequal, height=\plothalf,
    at={(return.south west)}, anchor=north west, yshift=-\plotsep,
    title={\textbf{(c)} Hours to return 0.82},
    xlabel={Imagination horizon $H$},
    xmode=log, xmin=6, xmax=170,
    xtick={8,16,32,64,128}, xticklabels={8,16,\textbf{32},64,128},
    minor xtick={7,10,12,14,20,24,28,40,48,56,80,96,112,160},
    ymin=0, ymax=5, ytick={0,2,4},
    legend style={at={(1,1)}, anchor=south east}, reverse legend,
]
\draw[highlight, line width=1em] (axis cs:32,0) -- (axis cs:32,5);
\addplot[ours, mark=*, mark options={fill=white}] coordinates {(32,1.37) (64,1.34) (128,3.80)};
\addlegendentry{no critic}
\addplot[ours, very thick, mark=*] coordinates {(16,1.19) (32,0.88) (64,1.70) (128,4.62)};
\addlegendentry{critic}
\addplot[ours, very thick, only marks, mark=x, mark size=1.8pt, forget plot] coordinates {(8,5)};
\addplot[ours, only marks, mark=x, mark size=1.8pt, forget plot] coordinates {(8,4.4)};
\addplot[ours, only marks, mark=x, mark size=1.8pt, forget plot] coordinates {(16,5)};
\node[context, anchor=north] at (axis cs:11.5,4.75) {not reached};
\end{axis}
\end{tikzpicture}
\caption{\textbf{In controlled comparisons, SuperMimic reaches a higher return sooner than PPO-Sim, and the critic enables shorter imagination horizons.} All runs use LAFAN1 (40 clips, \qty{2.45}{\hour}). (a)~SuperMimic backpropagates its imagined return through its world model on one GPU; PPO-Sim runs PPO in the simulator on 1, 4, or 8 GPUs (\S\ref{sec:scaling}). Each point on a curve is a saved checkpoint scored on all 40 clips. (b, c)~Final training return (b) and wall-clock hours until the training return reaches 0.82 (c) for each imagination horizon $H$, with and without the critic. The dashed line in (a) and (b) marks this target return of 0.82. Training return is the return of training episodes. Their initial states follow return-based reset sampling (Appendix~\ref{app:reset}) and shift during training, so training return differs from the checkpoint scores in (a). In (c), a cross marks a horizon at which return 0.82 is not reached: both variants at $H{=}8$, no critic at $H{=}16$. The shaded band marks the default $H{=}32$.}
\label{fig:controlled}
\end{figure}

Figure~\ref{fig:controlled}(a) gives a controlled comparison on LAFAN1~\citep{harvey2020robust}. The baseline, PPO-Sim, is PPO trained in the simulator. PPO-Sim shares SuperMimic's motions, observations, reward, termination, domain randomization, actor and critic network sizes, and evaluation protocol. It uses BeyondMimic's PPO hyperparameters~\citep{liao2025beyondmimic} and 8192 parallel environments per GPU, twice SuperMimic's. PPO-Sim estimates its policy gradient from sampled simulator rollouts; SuperMimic backpropagates its imagined return through world-model rollouts. SuperMimic reaches a return of 0.82 in evaluation within \qty{1.2}{\hour} on one GPU. Within the \qty{18}{\hour} budget, PPO-Sim does not reach it. Its final return is 0.60 on one GPU, 0.70 on four, and 0.72 on eight, so going from four to eight GPUs barely helps.

Figure~\ref{fig:compute}(b, c) shows that, for SuperMimic, GPU-hours to the target return grow slowly as dataset length increases, while on the same sweep the world model's prediction error stays stable; \S\ref{sec:wm-properties} traces both trends to one world model amortizing the dynamics shared across motions.

\subsection{Properties of the learned world model}
\label{sec:wm-properties}

These results rest on three properties of the learned world model $f_\omega$: its accuracy holds as the motion library grows, and its rollouts are fast and differentiable.

\noindent\textbf{One model amortizes the motion library.} In Figure~\ref{fig:compute}(c), the world model's 8-step prediction error stays between 0.007 and 0.011 as the dataset grows from 1 to \qty{264}{\hour}. Every motion is executed by the same humanoid, so added clips supply more evidence about the shared body and contact dynamics rather than demanding new capacity. This is consistent with the slow growth of the compute needed to reach the target return (\S\ref{sec:scaling}).

\begin{figure}[t]
\centering
\begin{tikzpicture}[font=\scriptsize, speedup/.style={context, |-|, shorten <=1ex, shorten >=1ex}]
\begin{axis}[
    name=speed, smplot, width=\plotequal, height=\plotshort,
    title={\textbf{(a)} Steps per second},
    xlabel={Parallel environments},
    xmode=log, xmin=400, xmax=84000,
    xtick={512,2048,8192,32768}, xticklabels={512,2k,8k,32k},
    minor xtick={1024,4096,16384,65536},
    ymode=log, ymin=1e4, ymax=3e7,
    ytick={1e4,1e5,1e6,1e7}, yticklabels={10\textsuperscript{4},10\textsuperscript{5},10\textsuperscript{6},10\textsuperscript{7}},
]
\fill[gray!12] (axis cs:2048,1e4) rectangle (axis cs:16384,3e7);
\node[font=\scriptsize, gray] at (axis cs:5793,1.1e7) {used in practice};
\draw[speedup] (axis cs:2048,7.84e4) -- node[left, black] {11$\times$} (axis cs:2048,8.43e5);
\draw[speedup] (axis cs:16384,2.61e5) -- node[right, black] {26$\times$} (axis cs:16384,6.71e6);
\draw[speedup] (axis cs:65536,3.22e5) -- node[left, black] {62$\times$} (axis cs:65536,2.00e7);
\addplot[ours, very thick, mark=*] coordinates {(512,2.16e5) (1024,4.22e5) (2048,8.43e5) (4096,1.68e6) (8192,3.35e6) (16384,6.71e6) (32768,1.34e7) (65536,2.00e7)} node[pos=0.25, above left] {World model};
\addplot[theirs, thick, mark=*, mark options={fill=white}] coordinates {(512,2.33e4) (1024,4.40e4) (2048,7.84e4) (4096,1.32e5) (8192,1.97e5) (16384,2.61e5) (32768,3.05e5) (65536,3.22e5)} node[pos=0.75, below right] {MuJoCo Warp};
\end{axis}
\begin{axis}[
    smplot, width=\plotequal, height=\plotshort,
    at={(speed.south east)}, anchor=south west, xshift=\plotgap,
    title={\textbf{(b)} Training return},
    xlabel={Imagined steps},
    xmode=log, xmin=1e7, xmax=8e10,
    table/x=plot_step, table/y=mean_reward_smooth,
    return axis,
    legend pos=south east, reverse legend, legend style={row sep=-3pt, name=ppolegend},
    legend image code/.code={\draw[#1] (0cm,0cm) -- (0.4cm,0cm);},
]
\draw[context, |-|] (axis cs:9.317e8,0.841) -- node[pos=0.2, below, inner sep=1pt, black] {64$\times$} (axis cs:5.956e10,0.841);
\addplot[theirslight!55!white, thick] table {figures/data/fig5b/walk_run_sprint_ppo.csv};
\addlegendentry{4k$\times$32}
\addplot[theirslight, thick] table {figures/data/fig5b/walk_run_sprint_ppo_b8192_n24.csv};
\addlegendentry{8k$\times$24}
\addplot[theirs, thick] table {figures/data/fig5b/walk_run_sprint_ppo_b8192.csv};
\addlegendentry{8k$\times$32}
\addplot[theirsdark, thick] table {figures/data/fig5b/walk_run_sprint_ppo_b16384.csv};
\addlegendentry{16k$\times$32}
\addplot[ours, very thick, forget plot] table {figures/data/fig5b/walk_run_sprint_supermimic.csv};
\node[anchor=east] at (axis cs:1.1e8,0.68) {SuperMimic};
\end{axis}
\node[anchor=south east, inner xsep=2pt] at (ppolegend.south west) {PPO-WM};
\end{tikzpicture}
\caption{\textbf{Imagined rollouts are fast and differentiable.} (a)~On an H100, over the batch sizes used in practice (shaded, 2k--16k parallel environments), the world model runs more than $10\times$ faster than MuJoCo Warp; the gap widens to $62\times$ at 64k, where MuJoCo Warp has flattened. One step is one \qty{50}{\hertz} control step, which MuJoCo Warp computes as four \qty{5}{\milli\second} substeps; the world model runs at the TF32 precision used in SuperMimic's training. (b)~On walk/run/sprint motions from LAFAN1~\citep{harvey2020robust}, SuperMimic, trained with its actor--critic update, reaches the best training return of PPO trained on world-model rollouts (PPO-WM) with about $64\times$ fewer imagined steps. PPO-WM runs at four batch sizes (parallel rollouts $\times$ steps per rollout); 16k$\times$32 is the best. Each curve is a centered moving average over $\pm0.05$ decade of imagined steps.}
\label{fig:wm-efficiency}
\end{figure}
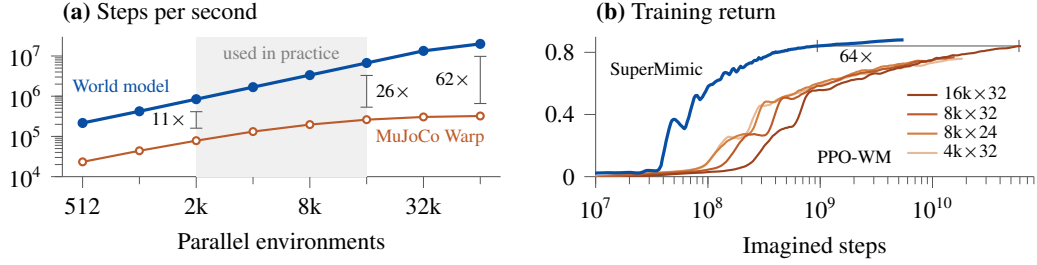

\noindent\textbf{Imagination accelerates rollouts.} During policy optimization, rollout speed sets the update cost. Figure~\ref{fig:wm-efficiency}(a) shows that the world model runs more than $10\times$ faster than MuJoCo Warp at the batch sizes used in practice, and up to $62\times$ at 64k parallel environments.

\noindent\textbf{First-order gradients need fewer imagined steps.} As a mechanism ablation, Figure~\ref{fig:wm-efficiency}(b) isolates the benefit of first-order policy updates: SuperMimic reaches the best PPO-WM return with about $64\times$ fewer imagined steps, so the gain is not just rollout speed.

\subsection{A critic enables shorter horizons and push recovery}
\label{sec:critic}

Figure~\ref{fig:controlled}(b, c) sweeps $H$ with and without the critic. Without the critic, training fails at $H\le16$. With it, $H{=}16$ matches the critic-free $H{=}32$, and $H{=}32$ reaches a return of 0.82 fastest. Raising $H$ from 64 to 128 nearly triples the time to that return and lowers the final return; $H{=}128$ is the longest horizon that fits on an H100. At matched return, the critic still changes behavior. Under \qtyrange{0.5}{4}{\metre\per\second} velocity pushes while holding a static T-pose in simulation, the critic policy settles within \qtyrange{0.5}{2.4}{\second}. The critic-free policy never settles: it keeps stumbling even at \qty{0.5}{\metre\per\second} and falls at \qty{4}{\metre\per\second}.
\subsection{Real-robot transfer}
\label{sec:real-robot}

\begin{figure}[t]
\centering
\newcommand{\snaprow}[3]{\begin{tikzpicture}[snaptag/.style={font=\footnotesize, text=white, fill=black, fill opacity=0.6, text opacity=1, inner sep=1.5pt, rounded corners=1pt}]
  \node[inner sep=0pt] (s) {\includegraphics[width=\linewidth]{#1}};
  \node[snaptag, anchor=north west, font=\footnotesize\bfseries] at ([shift={(1.5pt,-1.5pt)}]s.north west) {#2};
  \node[snaptag, anchor=north east] at ([shift={(-1.5pt,-1.5pt)}]s.north east) {$\Delta t=\qty{#3}{\second}$};
\end{tikzpicture}}
\snaprow{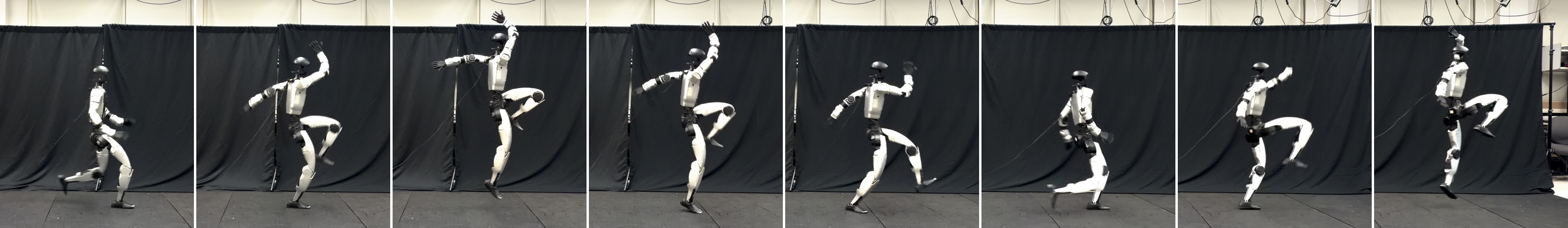}{(a)}{0.2}\\[0.5mm]
\snaprow{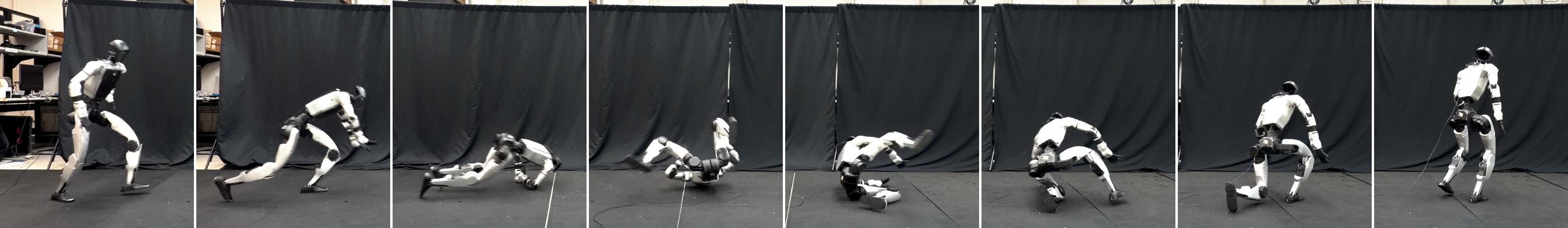}{(b)}{0.3}\\[0.5mm]
\snaprow{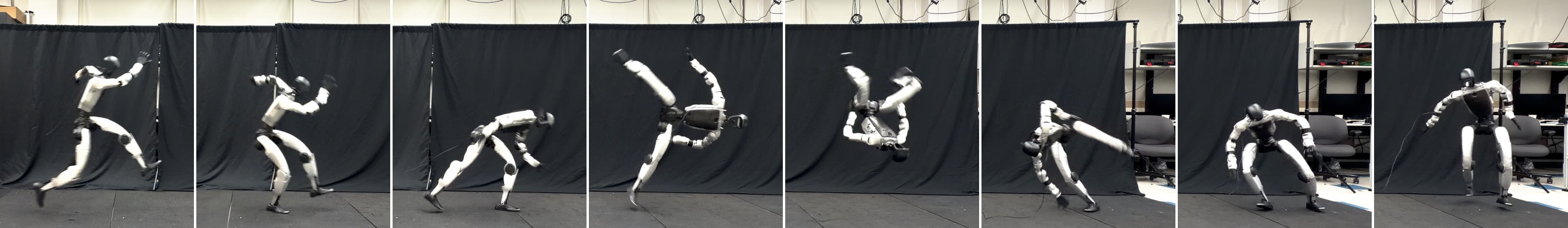}{(c)}{0.2}\\[0.5mm]
\snaprow{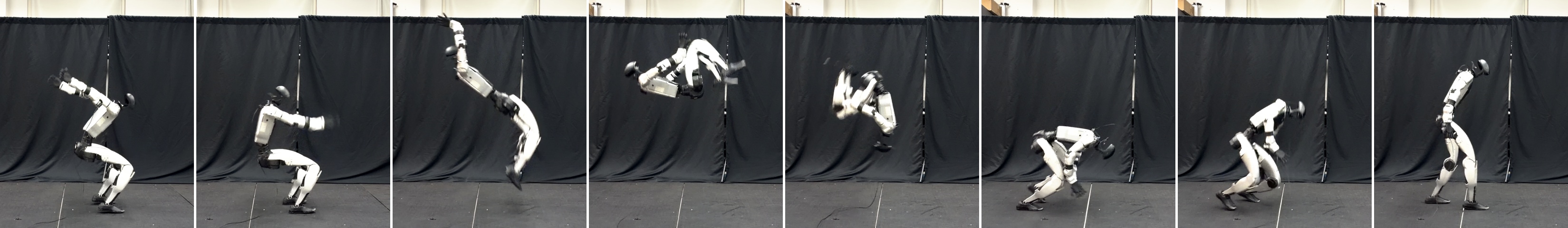}{(d)}{0.2}
\caption{\textbf{One general motion tracker, trained through the learned world model, transfers to a real Unitree~G1.} (a)~Skipping jump, (b)~side roll, (c)~$90^\circ$ flip, and (d)~in-place $360^\circ$ flip.}
\label{fig:acrobatics}
\end{figure}
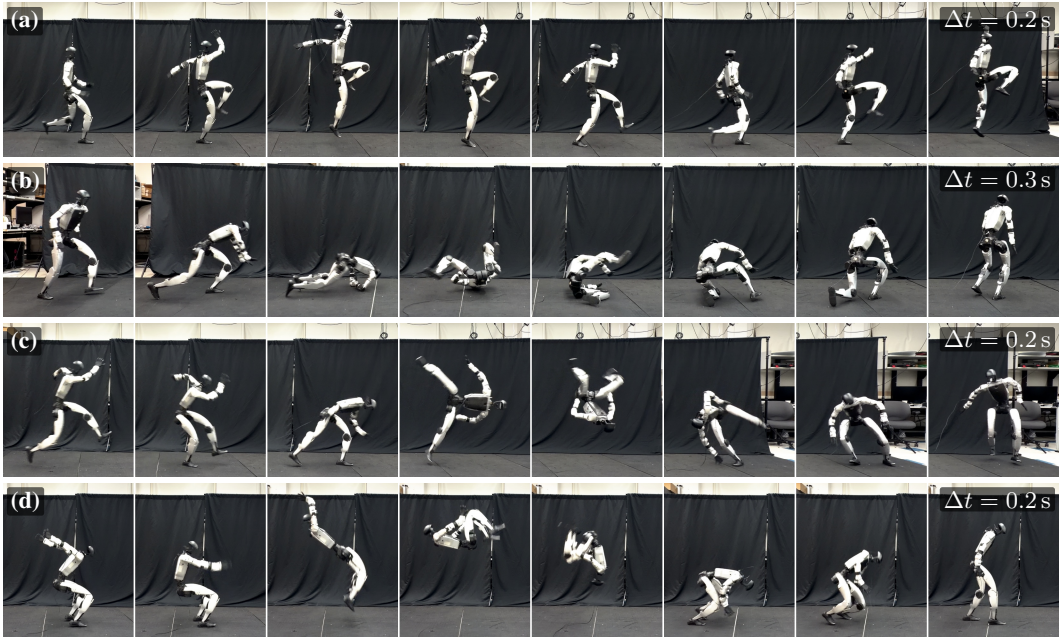

One general motion tracker deploys on a real Unitree~G1 without per-motion tuning (Figure~\ref{fig:acrobatics}). The four motions cover locomotion (a skipping jump), contact-rich whole-body motion with the torso and limbs on the floor (a side roll), and acrobatics (a $90^\circ$ flip and an in-place $360^\circ$ flip).

\subsection{Downstream tasks: RL encoder vs.\ distillation baseline}
\label{sec:downstream}

\begin{figure}[t]
\centering
\input{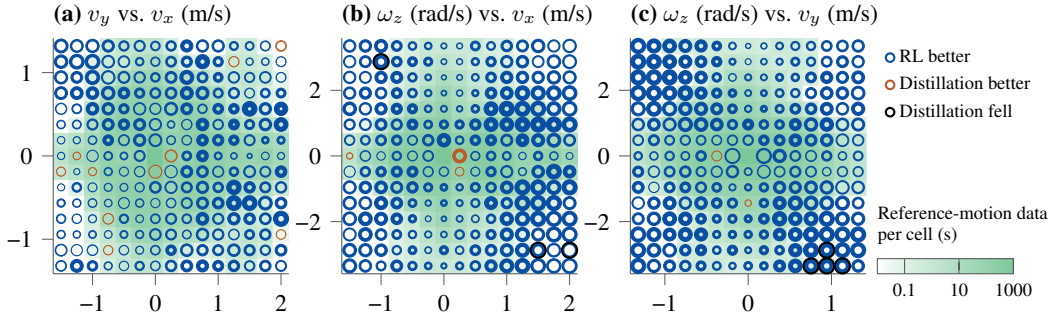}%
\pgfdeclareplotmark{ring}{%
  \pgfpathcircle{\pgfpointorigin}{\pgfplotmarksize}%
  \pgfpathcircle{\pgfpointorigin}{\stepgridhole}%
  \pgfseteorule\pgfusepath{fill}%
  \pgfpathcircle{\pgfpointorigin}{\pgfplotmarksize}%
  \pgfusepath{stroke}%
}
\newlength{\stepgridcell}\setlength{\stepgridcell}{\dimexpr\plotthird/\pgfkeysvalueof{/pgfplots/stepgrid cells}\relax}
\newcommand{\stepgridradius}[1]{0.48*\stepgridcell*sqrt(#1)}%
\pgfplotsset{
  stepgrid/.style={
    smplot, axis on top, width=\plotthird, height=\plotthird,
    colormap={stepgrid}{color(0cm)=(white) color(0.1cm)=(shade!15) color(1cm)=(shade)}, point meta min=-2, point meta max=3,
  },
  stepgrid field/.style={matrix plot*, point meta={log10(max(\thisrow{sec},0.01))}, mesh/cols=\pgfkeysvalueof{/pgfplots/stepgrid cells}},
  stepgrid outer/.style={visualization depends on={\thisrow{outer} \as \outer}, scatter/@pre marker code/.append style={/tikz/mark size={\stepgridradius{\outer}}}},
  stepgrid rings/.style={
    scatter, only marks, mark=ring, thin, scatter src=explicit symbolic, scatter/classes={rl={ours}, ds={theirs}},
    stepgrid outer, visualization depends on={\thisrow{inner} \as \inner},
    scatter/@pre marker code/.append code={\pgfmathsetlengthmacro{\stepgridhole}{\stepgridradius{\inner}}},
  },
  stepgrid fell/.style={scatter, only marks, mark=o, thick, scatter/use mapped color={draw=black}, stepgrid outer, restrict expr to domain={\thisrow{fell}}{1:1}},
}
\newcommand{\stepgridplots}[1]{%
  \addplot[stepgrid field] table[x=x, y=y, meta=sec] {figures/data/stepgrid_field_#1.csv};
  \addplot[stepgrid rings] table[x=x, y=y, meta=win] {figures/data/stepgrid_rings_#1.csv};
  \pgfplotstableread{figures/data/stepgrid_rings_#1.csv}\stepgridtable
  \def\stepgridfalls{0}
  \pgfplotstableforeachcolumnelement{fell}\of\stepgridtable\as\stepgridfell{\pgfmathtruncatemacro{\stepgridfalls}{\stepgridfalls+\stepgridfell}}
  \ifnum\stepgridfalls>0 \addplot[stepgrid fell] table[x=x, y=y] {\stepgridtable};\fi
}
\begin{tikzpicture}[font=\scriptsize]
\begin{axis}[name=a, stepgrid, stepgrid axis vx=x, stepgrid axis vy=y, title={\textbf{(a)} $v_y$ vs.\ $v_x$ (m/s)}]
\stepgridplots{vx_vy}
\end{axis}
\begin{axis}[name=b, stepgrid, stepgrid axis vx=x, stepgrid axis wz=y, at={(a.north east)}, anchor=north west, xshift=\plotthirdgap, title={\textbf{(b)} $\omega_z$ (rad/s) vs.\ $v_x$ (m/s)}]
\stepgridplots{vx_wz}
\end{axis}
\begin{axis}[
    name=c, stepgrid, stepgrid axis vy=x, stepgrid axis wz=y, at={(b.north east)}, anchor=north west, xshift=\plotthirdgap, title={\textbf{(c)} $\omega_z$ (rad/s) vs.\ $v_y$ (m/s)},
    legend style={at={(1,1)}, anchor=north west, xshift=1.5mm, font=\scriptsize},
    legend image code/.code={\draw[#1, mark size=1.6pt] plot coordinates {(0.12cm,0)};},
    colorbar horizontal, colorbar shift/.style={xshift=1.5mm},
    colorbar style={
      at={(parent axis.south east)}, anchor=south west, width=\dimexpr\plotlegendcol-3mm\relax, height=2mm,
      axis line style=context, tick style=context, tick label style={font=\scriptsize},
      xtick={-1,1,3}, xticklabels={0.1,10,1000},
      title={Reference-motion data\\per cell (s)}, title style={font=\scriptsize, at={(0,1)}, anchor=south west, inner xsep=0pt, align=left}, every axis title shift=2pt,
    },
]
\addlegendimage{ours, only marks, mark=o, thick}
\addlegendimage{theirs, only marks, mark=o, thick}
\addlegendimage{black, only marks, mark=o, thick}
\stepgridplots{vy_wz}
\legend{RL better, Distillation better, Distillation fell}
\end{axis}
\end{tikzpicture}
\caption{\textbf{The policy with the RL-trained velocity encoder tracks step commands more accurately than the distillation baseline.} Each panel is a plane of the command space with the third component at zero, and each ring is one command. The outer disc's area is proportional to the larger of the two policies' relative steady-state errors in the step response to that command, and the hole's area to the smaller one; the color marks the better policy. Details in Appendix~\ref{app:velocity}.}
\label{fig:downstream}
\end{figure}

SuperMimic trains downstream goal encoders jointly with the motion-tracking interface, sharing one decoder and world model (Appendix~\ref{app:unified}). Our study of downstream interfaces is preliminary. We train one velocity encoder on a velocity-tracking reward through the world model. We compare it with a distillation baseline: a velocity encoder trained on the motion-tracking reward, with commands read from the reference motions. Figure~\ref{fig:downstream} evaluates both on 672 step responses spanning the velocity space (Appendix~\ref{app:velocity}). The RL encoder has the lower steady-state error on 650 of them, and the distillation encoder on 15. Only the distillation encoder falls, on seven fast turns with almost no motion-library data.

\section{Conclusion}
SuperMimic removes compute as the bottleneck to scaling humanoid whole-body control. A single learned, differentiable world model captures the dynamics every motion shares, and backpropagation-through-time through it amortizes a whole dataset into one policy. The policy trains on a single GPU, surpasses the return of a policy trained on a 128-GPU cluster, and transfers to a real Unitree~G1. Its training cost grows slowly over a controlled dataset-length sweep. The same model can train a downstream velocity interface directly on the interface's own task reward. Measuring the scaling laws of motion tracking is now affordable, at one GPU per run.

\section{Limitations and Future Work}
\begin{itemize}[leftmargin=*,noitemsep,topsep=2pt]
  \item \textbf{Multi-interface training is understudied.} We train the velocity encoder concurrently with the motion tracker, but the velocity-tracking reward updates only the velocity encoder and not the shared decoder (Appendix~\ref{app:unified}). Concurrent training is therefore the same as training the velocity encoder after the motion tracker, and might not be necessary. Whether concurrent training helps once the velocity-tracking reward also updates the decoder is untested.
  \item \textbf{Simulator-only dynamics.} The world model learns only from simulator transitions. Real-robot data could refine it, and its differentiability could support system identification.
  \item \textbf{No terrain or object interaction.} SuperMimic trains only on flat ground. Extending it to terrain and object interaction is ongoing work.
\end{itemize}

\clearpage
\acknowledgments{We thank Bike Zhang, Sangli Teng, Nima Rahmanian, Songbo Hu, and Zhongyu Li for helpful discussions and help with the experiments. This work was supported in part by Komatsu, the BAIR Center for Humanoid Intelligence, Apple, and the RAI Institute. We thank Lambda for sponsoring compute for this project. Motion Data by Bones Studio (\url{https://bones.studio/}). Pieter Abbeel and Koushil Sreenath hold concurrent appointments as Professors at UC Berkeley and as Amazon Scholars. This paper describes work performed at UC Berkeley and is not associated with Amazon.}

\bibliography{root}

\clearpage
\appendix
\renewcommand{\thefigure}{A.\arabic{figure}}
\renewcommand{\thetable}{A.\arabic{table}}
\renewcommand{\theequation}{A.\arabic{equation}}
\renewcommand{\thealgorithm}{A.\arabic{algorithm}}
\renewcommand{\theHfigure}{A.\arabic{figure}}
\renewcommand{\theHtable}{A.\arabic{table}}
\renewcommand{\theHequation}{A.\arabic{equation}}
\setcounter{figure}{0}
\setcounter{table}{0}
\setcounter{equation}{0}
\setcounter{algorithm}{0}

\section{Instantiation: observation, action, reward, and sim context}
\label{app:spaces}

This appendix instantiates the shared spaces of the POMDP of \S\ref{sec:problem}. The motion-tracking interface follows the BeyondMimic formulation~\citep{liao2025beyondmimic}; the changes we make are noted inline.

\paragraph{Observation.}
The proprioceptive observation $\bm{o}_t$ comprises the joint positions relative to the default pose $\bm{q} - \bm{q}^0$, the joint velocities $\dot{\bm{q}}$, the base angular velocity from the inertial measurement unit (IMU), the projected gravity in the base frame, and the previous action $\bm{a}_{t-1}$ (Table~\ref{tab:obs-supermimic}); the policy receives the last 10 observations.

\begin{table}[!htbp]
\centering
\footnotesize
\setlength{\tabcolsep}{3pt}
\begin{tabular}{@{}llll@{}}
\toprule
\textbf{Observation Term} & \textbf{Dim.} & \textbf{Noise $\bm{\eta}_t$} & \textbf{Unit} \\
\midrule
Joint positions $\bm{q} - \bm{q}^0$ & 29 & $\mathcal{U}[-0.01,\,0.01]$ & \unit{\radian} \\
Joint velocities $\dot{\bm{q}}$ & 29 & $\mathcal{U}[-0.5,\,0.5]$ & \unit{\radian\per\second} \\
Base angular velocity (IMU) & 3 & $\mathcal{U}[-0.2,\,0.2]$ & \unit{\radian\per\second} \\
Projected gravity & 3 & $\mathcal{U}[-0.05,\,0.05]$ & -- \\
Previous action $\bm{a}_{t-1}$ & 29 & -- & -- \\
\bottomrule
\end{tabular}
\vspace{0.6em}
\caption{\textbf{Proprioceptive observation $\bm{o}_t$.} The noise column gives the per-dimension sensor noise $\bm{\eta}_t \sim \rho_\eta$.}
\label{tab:obs-supermimic}
\end{table}

\paragraph{Goal.}
For the motion-tracking interface, the goal $\bm{g}_t$ is read from the reference motion. 
Compared with BeyondMimic~\citep{liao2025beyondmimic}, we condition the encoder on a $K=10$ future reference sequence rather than a single-frame target. 
Other goal interfaces replace $\bm{g}_t$ with their corresponding command representation, such as target velocity commands (Appendix~\ref{app:unified}).

\paragraph{Action.}
The action $\bm{a}_t$ maps to a joint setpoint $\bm{q}^{\text{sp}} = \bm{\alpha}\,\bm{a}_t + \bm{q}^0$ about the default pose $\bm{q}^0$. Following BeyondMimic, the low PD gains make this setpoint a proxy for joint torque rather than a desired position.

\paragraph{Reward.}
We use BeyondMimic's motion-tracking reward (Table~\ref{tab:reward-supermimic}), removing its self-collision penalty. That penalty guards against reward hacking by a stochastic policy. When the reference brings two bodies close without contact, pressing them together suppresses the action noise and raises the velocity-tracking reward. Our policy is deterministic in imagined rollouts (\S\ref{sec:policy}), so the exploit has no payoff and the term is unnecessary. The joint-limit penalty is lowered from $-10$ to $-2$.

\paragraph{Termination.}
An episode terminates when the root body deviates from the reference beyond a height or tilt threshold ($|\bm{e}_{p,z}| > \qty{0.25}{\metre}$ or $e_{\mathrm{tilt}} > 0.8$). The tilt error $e_{\mathrm{tilt}}$ is the absolute difference between the $z$-components of the gravity direction expressed in the reference and robot root (pelvis) frames. For an upright reference, $e_{\mathrm{tilt}}$ reaches 0.8 at about $78^\circ$ of tilt. Unlike BeyondMimic, we do not terminate on end-effector deviation.

\paragraph{Sim context and domain randomization.}
The per-episode sim context $\bm{\xi}$ collects the domain-randomization parameters fed to the world model (\S\ref{sec:wm}); Table~\ref{tab:domain-rand-supermimic} lists them. We omit BeyondMimic's root-velocity perturbations: such pushes inject state discontinuities that the deterministic world model cannot capture, and we find randomization over the physical parameters alone suffices for transfer to hardware.

\begin{table}[t]
\centering
\begin{tabular}{llc}
\toprule
\textbf{Reward Term} & \textbf{Equation} & \textbf{Weight} \\
\midrule
\multicolumn{3}{l}{\emph{Task (tracking)}}\\
Body Position
& $\displaystyle
\exp\!\Big(
-\big( \tfrac{1}{|\mathcal{T}|}
\sum_{b\in\mathcal{T}}
\|\mathbf{p}^{\mathrm{des}}_{b}-\mathbf{p}_{b}\|^{2} \big) / 0.3^{2}
\Big)$
& $1.0$\\[2mm]
Body Orientation
& $\displaystyle
\exp\!\Big(
-\big( \tfrac{1}{|\mathcal{T}|}
\sum_{b\in\mathcal{T}}
\|\log(R^{\mathrm{des}}_{b}R_{b}^{\top})\|^{2} \big) / 0.4^{2}
\Big)$
& $1.0$\\[2mm]
Body Linear velocity
& $\displaystyle
\exp\!\Big(
-\big( \tfrac{1}{|\mathcal{T}|}
\sum_{b\in\mathcal{T}}
\|\mathbf{v}^{\mathrm{des}}_{b}-\mathbf{v}_{b}\|^{2} \big) / 1.0^{2}
\Big)$
& $1.0$\\[2mm]
Body Angular velocity
& $\displaystyle
\exp\!\Big(
-\big( \tfrac{1}{|\mathcal{T}|}
\sum_{b\in\mathcal{T}}
\|\boldsymbol{\omega}^{\mathrm{des}}_{b}-\boldsymbol{\omega}_{b}\|^{2} \big) / 3.14^{2}
\Big)$
& $1.0$\\[2mm]
Root Position
& $\displaystyle
\exp\!\Big(
-\|\mathbf{p}^{\mathrm{des}}_{\text{root}}-\mathbf{p}_{\text{root}}\|^{2} / 0.3^{2}
\Big)$
& $0.5$\\[2mm]
Root Orientation
& $\displaystyle
\exp\!\Big(
-\|\log(R^{\mathrm{des}}_{\text{root}}R_{\text{root}}^{\top})\|^{2} / 0.4^{2}
\Big)$
& $0.5$\\[2mm]
\midrule
\multicolumn{3}{l}{\emph{Regularization}}\\
Action smoothness
& $\|\mathbf{a}_{t}-\mathbf{a}_{t-1}\|^{2}$
& $-0.1$\\[2mm]
Joint limit
& $\sum_{j}\big[\max(0,\,q_{j}-q_{j}^{\max}) + \max(0,\,q_{j}^{\min}-q_{j})\big]$
& $-2$\\
\bottomrule
\end{tabular}
\vspace{0.6em}
\caption{\textbf{Motion-tracking interface reward, following BeyondMimic~\citep{liao2025beyondmimic} with the self-collision penalty removed and the joint-limit weight lowered from $-10$ to $-2$ (\S\ref{sec:policy}).} $\mathcal{T}$ is the set of tracked bodies.}
\label{tab:reward-supermimic}
\end{table}

\begin{table}[t]
\centering
\begin{tabular}{ll}
\toprule
\textbf{Domain Randomization} & \textbf{Sampling Distribution} \\
\midrule
Friction coefficients & $\mu \sim \mathcal{U}[0.3,\,1.2]$ \\
Joint calibration offsets [rad] & $ \Delta q^0_j \!\sim  \mathcal{U}[-0.01,\,0.01] $\\
Torso's center-of-mass offset [m]& $\Delta x\!\sim\!\mathcal{U}[-0.025,0.025]$,  $\ \Delta y\!\sim\!\mathcal{U}[-0.05,0.05]$, \\ &$ \Delta z\!\sim\!\mathcal{U}[-0.05,0.05]$ \\
\bottomrule
\end{tabular}
\vspace{0.6em}
\caption{\textbf{Domain-randomization parameters that make up the sim context $\bm{\xi}$, following BeyondMimic~\citep{liao2025beyondmimic} with the root-velocity perturbations removed.} ($\mathcal{U}[\cdot]$: uniform distribution.)}
\label{tab:domain-rand-supermimic}
\end{table}

\section{Joint training across goal interfaces}
\label{app:unified}
In \S\ref{sec:method}, the goal $\bm{g}_t$ is read from a full reference motion. Downstream goal interfaces specify the motion only in part or in another form. The same dynamics model can train several goal interfaces because the body physics is shared. We instantiate one POMDP $\mathcal{M}_i$ per goal interface (for example, a full-reference motion target, a velocity command, end-effector poses, or an SMPL body-pose target), each with its own goal space $\mathcal{G}_i$, reward $r_i$, and termination $d_i$, and train each with its own critic $V_{\psi_i}$. A single policy network uses per-interface goal encoders $\mathrm{enc}_{\theta,i}$ and a shared decoder $\mathrm{dec}_\theta$,
\begin{equation}
  \bm{z} = \mathrm{enc}_{\theta,i}(\bm{g}) \in \mathcal{Z}, \qquad \bm{a} = \mathrm{dec}_\theta(\bm{z}, \bm{o}), \qquad \bm{g} \in \mathcal{G}_i,
\end{equation}
and training optimizes the lifted objective $\max_\theta \mathbb{E}_{i\sim p(\mathcal{I})}[\hat{J}_i(\theta)]$, where $p(\mathcal{I})$ is a distribution over the set $\mathcal{I}$ of goal interfaces. Each interface $i$ thus has its own policy $\bm{a} = \pi_i(\bm{o}, \bm{g}) := \mathrm{dec}_\theta(\mathrm{enc}_{\theta,i}(\bm{g}), \bm{o})$; the velocity encoder also receives a feature of the observation history (Table~\ref{tab:policy-nets}). Each $\hat{J}_i$ is the objective of Eq.~\ref{eq:policy-obj} with $\pi_i$ in place of $\pi_\theta$, the reward and termination of $\mathcal{M}_i$, and the critic $V_{\psi_i}$. Every interface trains through the same dynamics model $f_\omega$, which is not retrained per interface.

A downstream interface is trained by distillation or by RL, depending on its reward. Distillation uses the motion-tracking reward (\S\ref{sec:policy}), with the command read from the reference motion. The reward gradient then trains that interface to reproduce the reference motion from this command. RL optimizes the interface's own task reward, such as velocity tracking, directly through the world model. On a downstream interface's rollouts, the return gradient passes through the shared decoder to that interface's encoder, with a stop-gradient on the decoder's weights; the decoder is still trained by the motion-tracking objective. The RL-trained velocity encoder of \S\ref{sec:downstream} also has a latent-alignment loss toward the motion encoder's latent on rollouts with reference commands (Appendix~\ref{app:velocity}).

\section{Network architectures}

Table~\ref{tab:policy-nets} lists the world-model dynamics network of \S\ref{sec:wm} and the policy and critic networks of Appendix~\ref{app:unified}.

\begin{table}[t]
\centering
\footnotesize
\setlength{\tabcolsep}{3pt}
\begin{tabular}{@{}>{\raggedright\arraybackslash}p{0.24\linewidth}>{\raggedright\arraybackslash}p{0.28\linewidth}>{\raggedright\arraybackslash}p{0.08\linewidth}>{\raggedright\arraybackslash}p{0.32\linewidth}@{}}
\toprule
\textbf{Network} & \textbf{Input} & \textbf{Output} & \textbf{Architecture} \\
\midrule
Dynamics $f_\omega$ & $\hat{\bm{s}}_t,\ \bm{a}_t,\ \bm{\xi}$ & $\hat{\bm{s}}_{t+1}$ & MLP (2048, 2048, 2048) \\
\midrule
Motion encoder $\mathrm{enc}_{\theta,\text{motion}}$ & $K{=}10$ future reference frames & $\bm{z}$ & MLP (2048, 1024, 512, 256) \\
Velocity encoder $\mathrm{enc}_{\theta,\text{vel}}$ (Appendix~\ref{app:velocity}) & normalized command $(v_x, v_y, \omega_z)$, feature of the last 10 frames of $\bm{o}$ & $\bm{z}$ & MLP (256, 256) \\
Shared decoder $\mathrm{dec}_\theta$ & $\bm{z}$, last 10 frames of $\bm{o}$ & $\bm{a}_t$ & MLP (2048, 1024, 512, 256) \\
Critic $V_{\psi_i}$ & $\hat{\bm{s}}_t,\ \bm{g}_t,\ \bm{\xi}$ & scalar & MLP (256, 128); one critic per interface \\
\bottomrule
\end{tabular}
\vspace{0.6em}
\caption{\textbf{World-model, policy, and critic networks.} MLP: multilayer perceptron, listed by hidden widths. The latent $\bm{z}$ is 32-dimensional and bounded by $\tanh$.}
\label{tab:policy-nets}
\end{table}

\section{Training loop and hyperparameters}
\label{app:algorithm}

\subsection{Algorithm}

Algorithm~\ref{alg:train} summarizes the SuperMimic training loop of \S\ref{sec:problem}; its three steps correspond one-to-one to the panels of Figure~\ref{fig:pipeline}.

\begin{algorithm}[t]
\caption{\textbf{SuperMimic training.} Both $\omega$ and $\theta$ update every iteration; $\omega$ is held fixed (stop-gradient) only inside the policy's backward pass. For a downstream interface, $\hat{J}_i$ updates only its encoder $\mathrm{enc}_{\theta,i}$.}
\label{alg:train}
\begin{algorithmic}[1]\small
\Require world model $f_\omega$; policy $\pi_\theta$ shared across interfaces $\{\mathcal{M}_i\}_{i\in\mathcal{I}}$; replay buffer $\mathcal{B}$
\For{each iteration}
    \State \(\triangleright\)~\textbf{Collect.}\ \ Roll out $\pi_\theta$ with action noise in simulation; push transitions to $\mathcal{B}$, replacing the oldest
    \State \quad Draw reset start points by return-based sampling \Comment{Appendix~\ref{app:reset}}
    \State \(\triangleright\)~\textbf{Fit world model.}\ \ Sample a logged $k$-step window from $\mathcal{B}$, roll $f_\omega$ out autoregressively on the logged actions, and step $\omega \gets \omega - \alpha_\omega \nabla_\omega \mathcal{L}_{\text{WM}}$ \Comment{Eq.~\ref{eq:wm-loss}}
    \State \(\triangleright\)~\textbf{Improve policy.}\ \ Sample an interface $i \sim p(\mathcal{I})$ and a start state from $\mathcal{B}$
    \State \quad Imagine $H$ steps on-policy through the fixed $f_\omega$:\ \ $\hat{\bm{s}}_{t+1} = f_\omega\big(\hat{\bm{s}}_t,\, \pi_i(\hat{\bar{\bm{o}}}_t+\bm{\eta}_t, \bm{g}_t),\, \bm{\xi}\big)$
    \State \quad Step $\theta \gets \theta + \alpha_\theta \nabla_\theta \hat{J}_i$ \Comment{Eq.~\ref{eq:policy-obj}; analytic gradients through $f_\omega$; critic $V_{\psi_i}$}
    \State \quad Fit the critic $V_{\psi_i}$ to the GAE targets of the imagined rollout \Comment{Eq.~\ref{eq:critic}}
\EndFor
\end{algorithmic}
\end{algorithm}

\paragraph{Return-based reset sampling.}
\label{app:reset}
Simulator episodes reset to points of the motion library, favoring those where the policy currently earns low return. The motion timeline is divided into \qty{1}{\second} bins, and each bin $b$ keeps a running estimate $G_b$ of the discounted return from it, updated on the policy's own rollouts. Resets mix three distributions: 50\% uniform over bins, 40\% from $\mathrm{softmax}_b(-25\,G_b/G_{\max})$, where $G_{\max}$ is the largest attainable discounted return, and 10\% in proportion to each bin's past softmax weight, so a hard motion keeps some practice after it is learned.

\subsection{Hyperparameters}

Table~\ref{tab:hyperparams} lists the hyperparameters. Where they differ, values are given for the one-GPU and the two-GPU run of Table~\ref{tab:failed-motion-stats}.

\begin{table}[p]
\small
\centering
\begin{tabular}{@{}ll>{\raggedright\arraybackslash}p{0.42\linewidth}@{}}
\toprule
\textbf{Hyperparameter} & \textbf{Symbol} & \textbf{Value} \\
\midrule
\multicolumn{3}{l}{\emph{World model}}\\
Window length & $k$ & 8 \\
Optimizer & & Adam \\
Learning rate & $\alpha_\omega$ & 1 GPU: \num{2e-4}; 2 GPUs: $\sqrt{2}\times\num{2e-4}$ \\
Batch size & & 10 batches of 2048 windows per update \\
Gradient clipping & & global norm 1.0 \\
\midrule
\multicolumn{3}{l}{\emph{Imagination and policy}}\\
Imagination horizon & $H$ & 32 \\
Optimizer & & Adam, $\beta = (0.9,\,0.95)$ \\
Learning rate & $\alpha_\theta$ & same as $\alpha_\omega$ \\
Batch size & & 4096 imagined rollouts per update \\
Gradient clipping & & global norm 1.0, then element-wise 0.1 \\
Discount & $\gamma$ & 0.99 \\
Sensor noise & $\bm{\eta}_t\sim\rho_\eta$ & Table~\ref{tab:obs-supermimic} \\
\midrule
\multicolumn{3}{l}{\emph{Critic}}\\
GAE parameter & $\lambda$ & 0.95 \\
Optimizer & & Adam \\
Learning rate & & same as $\alpha_\omega$ \\
Gradient clipping & & global norm 1.0 \\
Critic epochs per update & & 5, each over 4 shuffled minibatches \\
\midrule
\multicolumn{3}{l}{\emph{Data collection}}\\
Simulator environments per GPU & & \num{4096} \\
Action noise & $\bm{\epsilon}_t\sim\rho_\epsilon$ & $\mathcal{N}(0,\,0.1^2)$ per action dimension \\
Replay-buffer size & & 1000 steps per environment \\
\midrule
\multicolumn{3}{l}{\emph{Schedule}}\\
Update-to-data ratio & & 0.4 \\
Policy updates per world-model update & & 1 \\
Learning-rate schedule & & cosine decay to $0.03\times$ over \num{250000} iterations, then constant \\
Training length & & \num{320000} iterations \\
Multi-GPU & & data parallel; gradients averaged across GPUs \\
\bottomrule
\end{tabular}
\vspace{0.6em}
\caption{\textbf{SuperMimic hyperparameters.}}
\label{tab:hyperparams}
\vspace{1.2em}
\centering
\begin{tabular}{llc}
\toprule
\textbf{Reward Term} & \textbf{Equation} & \textbf{Weight} \\
\midrule
\multicolumn{3}{l}{\emph{Task (velocity tracking)}}\\
Linear velocity
& $\displaystyle \exp\!\Big( -\|\mathbf{v}_{xy}-\mathbf{v}^{\mathrm{cmd}}_{xy}\|^{2} / 0.5^{2} \Big)$
& $2.0$\\[1mm]
Yaw rate
& $\displaystyle \exp\!\Big( -(\omega_{z}-\omega^{\mathrm{cmd}}_{z})^{2} / 0.5 \Big)$
& $2.0$\\[1mm]
\midrule
\multicolumn{3}{l}{\emph{Regularization}}\\
Action smoothness
& $\|\mathbf{a}_{t}-\mathbf{a}_{t-1}\|^{2}$
& $-0.3$\\[1mm]
Joint limit
& $\sum_{j}\big[\max(0,\,q_{j}-q_{j}^{\max}) + \max(0,\,q_{j}^{\min}-q_{j})\big]$
& $-2$\\[1mm]
Torso sideways lean
& $\sin^{2}\phi_{\mathrm{torso}}$
& $-4$\\[1mm]
Torso roll and pitch rate
& $\omega_{\mathrm{roll}}^{2}+\omega_{\mathrm{pitch}}^{2}$
& $-0.05$\\
\midrule
\multicolumn{3}{l}{\emph{Termination (fall)}}\\
\multicolumn{3}{l}{$h_{\mathrm{root}} < \qty{0.35}{\metre}$ \;or\; $1 + u_{z} > 0.7$}\\
\bottomrule
\end{tabular}
\vspace{0.6em}
\caption{\textbf{Velocity-tracking reward and termination used to train the RL encoder.} $\mathbf{v}_{xy}$ is the planar root velocity in the heading frame, $\omega_z$ the world yaw rate, and $(\mathbf{v}^{\mathrm{cmd}}_{xy}, \omega^{\mathrm{cmd}}_{z})$ the command. $\phi_{\mathrm{torso}}$ is the sideways lean of the torso $z$-axis (forward lean is not penalized) and $\omega_{\mathrm{roll}}, \omega_{\mathrm{pitch}}$ the torso roll and pitch rates, all in the heading frame. $h_{\mathrm{root}}$ is the root height and $u_z$ the $z$-component of the unit gravity direction in the root frame, so $1+u_z$ is 0 upright and 2 upside down, and the tilt condition triggers at about $73^\circ$. Every term, here and in Table~\ref{tab:reward-supermimic}, is multiplied by the control period $\Delta t=\qty{0.02}{\second}$.}
\label{tab:reward-velocity}
\end{table}

\section{Motion data filtering}
\label{app:data-filter}

We train on the publicly available G1-retargeted BONES-SEED motion set~\citep{bones2026seed}. The raw
release contains \num{142220} clips, but many interact with terrain or objects the
robot cannot use (stairs, boxes, ladders, chairs) or carry retargeting artifacts
that make them infeasible for a free-standing humanoid on flat ground. We remove these
in two passes and keep \num{129514} clips (Table~\ref{tab:data-filter}).

\paragraph{Name filter.}
Each clip is named by its action. We drop any clip whose name contains a term
denoting contact with terrain or an object the robot cannot use---e.g.\
\texttt{stair}, \texttt{ladder}, \texttt{box}, \texttt{chair}, \texttt{sit},
\texttt{climb}, \texttt{jump\_off\_50cm}, \texttt{look\_over\_fence}. This pass
(adapted from the keyword list in SONIC's released data-processing
code~\citep{luo2025sonic}, plus a few terms it misses) removes
\num{12553} clips.

\paragraph{Feasibility filter.}
The name filter misses clips whose names do not reveal the problem (e.g.\ a
routine that walks onto a box). We run each motion through MuJoCo forward
kinematics and flag it from the resulting body-link and foot-geometry trajectories, writing
$z_{\min}(t)=\min_b z_b(t)$ for the height of the lowest body link at time $t$ with the floor at
$z=0$:
\begin{itemize}
  \item \textbf{Elevated support.} $z_{\min}$ holds above \qty{0.20}{\metre} for at
  least \qty{0.6}{\second} (standing or climbing on a box).
  \item \textbf{Ground penetration.} $z_{\min}$ falls below \qty{-0.12}{\metre}, or
  below \qty{-0.08}{\metre} for over 80\% of the clip (legs driven through the
  floor by a faulty retarget).
  \item \textbf{Floating feet.} For at least \qty{0.5}{\second} \emph{both} feet
  stay more than \qty{0.02}{\metre} above the floor---measured as MuJoCo's exact
  geom-to-plane distance (\texttt{mj\_geomDistance}) from the foot collision
  geometry, not a link-origin proxy---while the pelvis is upright and at standing
  height and the body is nearly stationary. A free-standing humanoid cannot hold
  both feet off the floor while standing still, so this is infeasible on flat
  ground whatever the cause (a low platform, a retarget float, or a slow rise; a
  ballistic jump is excluded by the stationary gate). The clean ground stance
  sits at \qty{-0.004 \pm 0.004}{\metre}, so the \qty{0.02}{\metre} gate is several
  $\sigma$ clear of the floor.
\end{itemize}
This pass removes a further 153 clips. All thresholds are tuned for the G1
retarget at $50\,\mathrm{Hz}$.

\begin{table}[t]
\centering
\begin{tabular}{lr}
\toprule
\textbf{Stage} & \textbf{Clips} \\
\midrule
Raw BONES-SEED (G1 retarget) & \num{142220} \\
After name filter & \num{129667} \\
After feasibility filter & \num{129514} \\
\bottomrule
\end{tabular}
\vspace{0.6em}
\caption{\textbf{Number of BONES-SEED clips before filtering and after each filter.}}
\label{tab:data-filter}
\end{table}

\section{Evaluation}
\label{app:eval}
\subsection{Benchmark metrics}

We evaluate SONIC and SuperMimic on the same filtered BONES-SEED training-set
benchmark, containing \num{129514} evaluated motions. For SONIC we evaluate its
released checkpoint and its v1.1
checkpoint~\citep{luo2025sonic}. Each motion plays once from its first frame,
with the same simulation for every policy: foot friction 1.0, no domain
randomization or observation noise, and deterministic physics. Each policy uses
its own actuator gains and action scale, and receives its observations as its
training environment builds them. A motion is marked as failed if the rollout
triggers SuperMimic's training termination (Appendix~\ref{app:spaces}). SONIC's own success criterion
fails a motion on root and end-effector height deviations above \qty{0.25}{\metre}, without
an orientation term~\citep{luo2025sonic}.

\subsection{Failed motions}

\begin{figure}[H]
\centering
\begin{tikzpicture}[font=\scriptsize]
\begin{axis}[
    smplot, width=\plotfull, height=\plottall,
    title={Failed motions},
    xlabel={Wall-clock time (h)},
    xmin=0, xmax=10, ymin=0, ymax=500,
    legend pos=north east,
]
\addplot[theirs, densely dashed, forget plot] coordinates {(0,179) (10,179)} node[pos=1, left, fill=white, inner sep=2pt] {SONIC checkpoint, 179};
\addplot[ours, very thick] coordinates {
      (0.12,1554) (0.24,646) (0.36,455) (0.47,393) (0.59,216) (0.71,172) (0.82,159) (0.94,148) (1.06,105)
      (1.18,88) (1.29,103) (1.41,83) (1.53,61) (1.64,89) (1.76,75) (1.88,49) (2.00,49) (2.11,48) (2.23,47)
      (2.35,68) (2.46,30) (2.58,37) (2.70,28) (2.81,28) (2.93,27) (3.05,31) (3.16,30) (3.28,24) (3.40,24)
      (3.51,16) (3.63,18) (3.75,28) (3.87,21) (3.98,23) (4.10,18) (4.22,14) (4.33,14) (4.45,12) (4.57,15)
      (4.68,22) (4.80,14) (4.92,9) (5.04,13) (5.15,10) (5.27,7) (5.39,8) (5.50,6) (5.62,13) (5.74,7)
      (5.85,9) (5.97,8) (6.09,5) (6.21,9) (6.32,9) (6.44,9) (6.56,9) (6.67,9) (6.79,11) (6.91,6)
      (7.02,9) (7.14,7) (7.26,8) (7.38,7) (7.49,2) (7.61,5) (7.73,8) (7.84,7) (7.96,8) (8.08,4)
      (8.20,4) (8.31,8) (8.43,4) (8.55,4) (8.66,5) (8.78,5) (8.90,5) (9.02,3) (9.13,7) (9.25,3)
      (9.37,6)
    };
\addlegendentry{SuperMimic, 1 H100 GPU}
\addplot[ours!50, very thick] coordinates {
      (0.13,899) (0.25,273) (0.37,143) (0.49,162) (0.61,93) (0.74,84) (0.86,55) (0.98,74) (1.10,46)
      (1.23,55) (1.35,45) (1.47,48) (1.59,59) (1.72,51) (1.84,42) (1.96,42) (2.08,25) (2.21,28) (2.33,25)
      (2.45,26) (2.57,33) (2.69,29) (2.82,28) (2.94,21) (3.06,18) (3.18,16) (3.31,14) (3.43,11) (3.55,14)
      (3.67,14) (3.79,8) (3.92,8) (4.04,8) (4.16,5) (4.28,6) (4.41,8) (4.53,5) (4.65,2) (4.77,4)
      (4.90,3) (5.02,6) (5.14,2) (5.26,4) (5.39,1) (5.51,2) (5.63,2) (5.75,1) (5.87,4) (6.00,1)
      (6.12,0) (6.24,0) (6.36,1) (6.49,0) (6.61,0) (6.73,0) (6.85,0) (6.98,0) (7.10,0) (7.22,0)
      (7.34,0) (7.47,0) (7.59,1) (7.71,0) (7.83,0) (7.96,0) (8.08,1) (8.20,0) (8.32,0) (8.45,0)
      (8.57,0) (8.69,0) (8.81,0) (8.93,0) (9.06,0) (9.18,0) (9.30,0) (9.42,0) (9.55,0) (9.67,0)
      (9.79,0)
    };
\addlegendentry{SuperMimic, 2 H100 GPUs}
\end{axis}
\end{tikzpicture}
\caption{\textbf{Failed motions during training on the filtered BONES-SEED training-set benchmark.} Checkpoints are scored every \num{4000} iterations, for the runs of Figure~\ref{fig:compute}(a). SONIC: released checkpoint.}
\label{fig:failed-motions}
\end{figure}
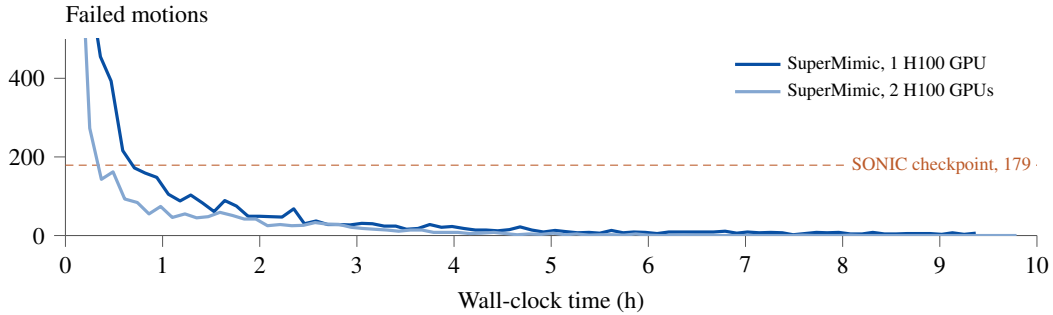

Table~\ref{tab:failed-motion-stats} counts failed motions per motion group. SONIC fails whole groups (all 44 flips, all 22 side rolls, and all 8 motions of one dance routine) and 15 of the 24 broad jumps. Its v1.1 checkpoint fails 239 motions, among them the same flips and dance motions and 19 of the 22 side rolls. After \qty{3}{\hour} on one GPU, SuperMimic's failures concentrate in a few small groups of hard motions: the first five groups of Table~\ref{tab:failed-motion-stats}, 195 motions, hold 14 of its 25 failures. In no group does it fail more motions than SONIC. After \qty{9.8}{\hour} on two GPUs, SuperMimic fails on none. Figure~\ref{fig:failed-motions} shows the count over training. The benchmark motions are training motions for both methods, so the counts measure how fully each policy covers these training motions, not generalization to new motions.

\begin{table}[H]
\centering
\footnotesize
\begin{tabular}{@{}lrrr@{}}
\toprule
 & & \multicolumn{2}{c}{\textbf{SuperMimic}} \\
\cmidrule(l){3-4}
\textbf{Motion group (motions)} & \textbf{SONIC} & \textbf{1 GPU $\times$ \qty{3}{\hour}} & \textbf{2 GPUs $\times$ \qty{9.8}{\hour}} \\
\midrule
side rolls (22) & \textbf{22} & 2 & 0 \\
flips: $360^\circ$, $180^\circ$, $90^\circ$ (44) & \textbf{44} & 1 & 0 \\
fire-in-the-hole turns (20) & 3 & 3 & 0 \\
crossed legs (10) & 9 & 0 & 0 \\
lying, resting and standing up (99) & 28 & 8 & 0 \\
\midrule
broad jumps (24) & 15 & 0 & 0 \\
dance routine (8) & \textbf{8} & 0 & 0 \\
all other motions (\num{129287}) & 50 & 11 & 0 \\
\midrule
total (\num{129514}) & 179 & 25 & 0 \\
\bottomrule
\end{tabular}
\vspace{0.6em}
\caption{\textbf{Failed motions per motion group on the filtered BONES-SEED training-set benchmark.} SONIC: released checkpoint. Bold: the policy fails every motion in the group.}
\label{tab:failed-motion-stats}
\end{table}

\subsection{Experiment protocols}

\paragraph{Figure~\ref{fig:compute}.}
SuperMimic trains on the filtered BONES-SEED set (Appendix~\ref{app:data-filter}). The dataset-length sweep in (b, c) uses subsets of 1, 7, 40, and \qty{264}{\hour}.

\section{Velocity-command task}
\label{app:velocity}

\subsection{Training setup}
The velocity encoder of \S\ref{sec:downstream} maps the command $(v_x, v_y, \omega_z)$ and a feature of the last 10 frames of $\bm{o}$ to the shared latent.

Each encoder is trained from scratch in its own run, jointly with the motion-tracking interface, for \num{100000} iterations. Both runs use a walking and running library filtered from LAFAN1 and BONES-SEED. Half of the 4096 simulator environments and half of each imagined batch use the velocity interface, so $p(\mathcal{I})$ is uniform; unlike Algorithm~\ref{alg:train}, which samples one interface per update, each imagined batch mixes both.

\noindent\textbf{Commands.} The command $\bm{g}_t = (v_x, v_y, \omega_z)$ stacks the target planar root velocity in the heading (yaw-only) frame and the target world-frame yaw rate. A reference command is the centered 25-frame mean of the reference root velocity in the reference's heading frame. For the RL encoder, each environment resamples its command source at every reset and every 250 control steps (\qty{5}{\second}): with probability 0.5 it reads the command from the reference clip at every step, and otherwise it holds a command drawn uniformly from $v_x \in [\num{-1.5}, 2]$\,\unit{\metre\per\second}, $v_y \in [\num{-1.32}, 1.32]$\,\unit{\metre\per\second} and $\omega_z \in [\num{-3.35}, 3.35]$\,\unit{\radian\per\second}, zeroed (standing) with probability 0.1.

\noindent\textbf{Encoder.} On rollouts with reference commands, a mean-squared loss of weight 0.1 pulls the velocity latent toward the motion encoder's latent at the same step, with a stop-gradient on the motion latent.

\noindent\textbf{Rollouts, reward, and termination.} Each interface's imagined rollouts start from states logged by its own simulator environments. Velocity rollouts of the RL encoder use the velocity-tracking reward and a fall-only termination (Table~\ref{tab:reward-velocity}); those of the distillation baseline use the motion-tracking reward and termination of Appendix~\ref{app:spaces}.

\subsection{Step test}

\noindent\textbf{Protocol.} From standing, the policy receives a fixed command for \qty{5}{\second}; each command is tested in four trials. The relative steady-state error is the norm of the mean velocity error over the last \qty{2}{\second}, divided by the norm of the command, with the root velocity measured in the heading frame. The per-command value is the median over the trials that did not fall. The 672 step responses form a $15\times15$ grid on each of three planes, spanning the uniform command range above, with the standing command left out. In Figure~\ref{fig:downstream}, errors above 0.9 are drawn at 0.9, and a policy that fell in all four trials is drawn at the cap.

\noindent\textbf{Results.} The RL encoder has the lower error on 650 of the 672 step responses, and the distillation encoder on 15. Their median relative errors are 0.16 and 0.49. Distillation falls in every trial on seven commands, all combining translation with a turn of at least \qty{2.8}{\radian\per\second}, where the motion library has almost no data. Each policy has one training seed.

\end{document}